\documentclass[12pt]{article}

\usepackage{subcaption}
\usepackage[bottom]{footmisc}
\usepackage{indentfirst} 
\usepackage{stmaryrd}
\usepackage{endnotes}
\usepackage{verbatim}
\usepackage{tikz}
\usepackage{threeparttable}
\usepackage{comment}
\usepackage{algorithm}
\usepackage{titlesec}
\usepackage{algorithmic}
\usepackage{multirow}
\usepackage{makecell}
\usepackage{chngcntr}
\usepackage{dsfont}
\usepackage{enumitem}
\usepackage{graphicx,float}
\usepackage{amsmath,amsthm,amssymb,color}
\usepackage[utf8]{inputenc} %
\usepackage[T1]{fontenc}    %
\usepackage[round]{natbib}
\usepackage{hyperref}       %
\hypersetup{colorlinks=true, linkcolor=blue, citecolor=blue, urlcolor=blue}
\usepackage{url}            %
\usepackage{booktabs}       %
\usepackage{amsfonts}       %
\usepackage{nicefrac}       %
\usepackage{microtype}      %
\usepackage{xcolor}         %
\usepackage{algorithm}
\usepackage{titlesec}
\usepackage{algorithmic}
\usepackage{setspace}
\usepackage{array}
\usepackage{geometry}
\usepackage[bottom]{footmisc}

\newtheorem{assumpA}{Assumption}

\newcounter{condition}
\newcommand{\condlabel}[1]{%
  \refstepcounter{condition}\label{#1}%
}

\newcommand{\bS}{\mathrm{\bf S}}

\def\calD{\mathcal{D}}
\def\calE{\mathcal{E}}

\def\calP{\mathcal{P}}

\def\calS{\mathcal{S}}
\def\calT{\mathcal{T}}

\def\calX{\mathcal{X}}
\def\calY{\mathcal{Y}}

\newcommand{\bmu}{\boldsymbol{\mu}}

\newcommand{\bSigma}{\boldsymbol{\Sigma}}

\newcommand{\bzero}{\mathrm{\bf 0}}

\newcommand{\bbP}{\mathbb{P}}
\makeatletter
\@ifundefined{theorem}{\newtheorem{theorem}{Theorem}}{}
\@ifundefined{proposition}{\newtheorem{proposition}[theorem]{Proposition}}{}
\theoremstyle{definition}
\@ifundefined{definition}{\newtheorem{definition}[theorem]{Definition}}{}
\@ifundefined{remark}{\newtheorem{remark}[theorem]{Remark}}{}
\theoremstyle{plain}
\makeatother

\begin{document}

\title{\bf From Imbalance to Evidence: \\ Deep Nonparametric Testing for Complex Data \\ under Severe Sample Imbalance}
\author{Guanhao Zhou and Xiufan Yu\\[1ex]
{Department of Applied and Computational Mathematics and Statistics} \\[1ex]
{University of Notre Dame}}
\date{}
\maketitle

\setstretch{1.2}
\begin{abstract}
Comparing distributions under severe sample imbalance requires effectively using abundant majority observations while retaining sensitivity to differences involving the smaller sample. This challenge is particularly relevant for complex, non-Euclidean data, such as graphs, whose structural characteristics call for flexible testing procedures. Classifier two-sample tests (C2STs) provide such flexibility, but existing procedures are primarily designed for balanced or comparably sized samples. In this paper, we study C2STs in a highly imbalanced regime, and show that severe sample imbalance may substantially degrade testing performance. Different from conventional imbalanced learning approaches that treat sample imbalance merely as a nuisance for classifier training, we turn it into a source of repeated testing evidence. Building on this insight, we propose an Enhanced Imbalance-Aware Deep Classifier Two-Sample Test (EIAD-C2ST), a classifier-agnostic framework designed for sample imbalance that accommodates diverse data modalities by tailoring classifiers to data structure. For computationally expensive deep classifiers, we further develop a scalable representation-amortized implementation strategy. Theoretically, we establish rigorous guarantees on asymptotic size and characterize conditions for power consistency, allowing the imbalance ratio to diverge.  We demonstrate the empirical gains using simulations and an application to blockchain forensics. The results reveal that the majority-sample surplus, when properly leveraged, can be organized into useful statistical evidence, rather than an obstacle, for distributional comparisons with complex data.
\end{abstract}

\vspace{1ex}
\newpage

\setstretch{1.5}
\section{Introduction}

Many consequential real-world problems exhibit severe class imbalance, and the rare observations are often those of greatest interest. This imbalance arises naturally in many scientific and societal processes, where outcomes or events may be inherently rare or disproportionately distributed \citep{chen2024survey}. For instance, in fraud detection, fraudulent transactions are significantly fewer compared to legitimate activities  \citep{dalpozzolo2014lessons,makki2019experimental}; in medical studies, especially those involving rare diseases or case-control designs, recruiting participants with a specific disease may be difficult due to low prevalence, cost constraints, or ethical considerations  \citep{khushi2021comparative,wang2022imbalanced}.
Related concerns arise in LLM safety and content moderation, where harmful, toxic, or adversarial jailbreak prompts represent only a small fraction of the vast volume of queries that deployed language models receive daily, yet their detection is critical for safe and responsible deployment \citep{gehman2020realtoxicityprompts,shen2024anything}.  In these settings, balanced data collection may be infeasible. 
Beyond predicting individual labels, an important inferential question is whether the observed characteristics follow the same distribution across the two populations. 
This question becomes particularly challenging when one population is scarcely represented, even if the pooled dataset is large.

Two-sample testing provides the statistical framework for this distributional comparison. Given two independent samples, the goal is to determine whether they are drawn from the same underlying distribution. Classical methods, including many recent high-dimensional procedures, often impose assumptions about the data-generating distributions, the geometry of the sample space, or the relationship between sample size and dimension \citep{lehmann2005testing,huang2022overview}.
Most existing methods are primarily designed for structured Euclidean data. 
Modern datasets, however, increasingly involve complex data collected from irregular or complex domains (e.g., images, graphs, text, and audio), for which the structure of the observations is central to an informative comparison. When these data are also severely imbalanced, the methodological challenge is twofold: the test must accommodate the observation structure and use highly unequal samples effectively.
This intersection of sample imbalance and structural complexity motivates a pressing need for flexible nonparametric testing procedures that require minimal distributional assumptions and can operate beyond conventional Euclidean data.

A related but distinct literature studies imbalanced learning and focuses on approaches to improving
classifier performance \citep{he2009learning,krawczyk2016learning}. 
Representative approaches include resampling techniques \citep{chawla2002smote,liu2008exploratory,lemaitre2017imbalanced}, cost-sensitive learning \citep{elkan2001foundations,zhou2006training}, threshold adjustment \citep{sheng2006thresholding,zou2016finding}, and specialized loss functions \citep{lin2017focal,cui2019class}.
However, some of these remedies do not transfer directly to complex, non-Euclidean data.  A prominent example is the Synthetic Minority Oversampling Technique (SMOTE) \citep{chawla2002smote}, which faces several important challenges when applied to complex or high-dimensional data \citep{fernandez2018smote}.
In particular, SMOTE generates synthetic minority samples by interpolating between nearest neighbors in a feature space.
This construction implicitly assumes that Euclidean interpolation is meaningful, making it less suitable for non-Euclidean data such as graphs, text, audio, or other structured objects. 
In high-dimensional settings, SMOTE is further hampered by ``the curse of dimensionality'', where the concentration of distances erodes the meaningfulness of nearest-neighbor relationships \citep{blagus2013smote}.
Thus, complex data motivate imbalance-handling strategies that do not require synthesizing new observations in the original sample space.
Furthermore, this literature primarily focuses on improving classifier fitting. Such predictive objectives, however, do not by themselves solve the inferential problem considered here. 

For distributional inference, kernel-based methods provide an established class of flexible nonparametric tests, among which the Maximum Mean Discrepancy (MMD) test \citep{gretton2009fast,gretton2012kernel,schrab2023mmd} is one of the most prominent. MMD measures distributional discrepancy by embedding probability distributions into a reproducing kernel Hilbert space and computing the distance between their mean embeddings.
Although MMD is flexible and widely applicable, its empirical performance can be sensitive to kernel choice and hyperparameter tuning \citep{muandet2017kernel,ramdas2017wasserstein}. In recent years, \emph{Classifier Two-Sample Tests} (C2STs) \citep{lopez2017revisiting,kim2021classification} have emerged as a powerful alternative and have gained increasing attention in the machine learning community. C2ST recasts two-sample testing as a classification problem: under the null hypothesis that the two distributions are identical, a well-calibrated classifier should perform no better than chance; under the alternative, systematic distributional differences provide predictive signals that enable the classifier to distinguish between the two samples. 
This formulation makes C2ST especially attractive for modern applications as it can leverage flexible machine-learning classifiers,  including deep architectures, tailored to the data type.

Despite this flexibility, severe sample imbalance creates a challenge for C2ST that is fundamentally statistical rather than merely computational. Though the impact of class imbalance in classifier training can often be alleviated through techniques such as resampling, class weighting, threshold adjustment, or specialized loss functions, these strategies do not directly address the inferential goals and calibration requirements of two-sample testing. 
In C2ST, the goal is not only to train a classifier but also to construct a valid test statistic with proper calibration under the null.
A classifier trained on highly imbalanced pooled data may allocate most of its capacity to the majority sample, weakening sensitivity to distributional departures. 
Moreover, raw classification accuracy can be misleading: a classifier may appear to perform well by primarily fitting the majority class, while failing to capture the distributional differences most relevant for testing. Empirically, we observe that the performance of C2ST can be substantially affected by the sample-size ratio and deteriorates in highly imbalanced situations; see Figure~\ref{fig:plot_asymp_dist} for empirical evidence. %

The severely imbalanced regime also calls for theoretical analysis beyond conventional proportional-sample-size settings. 
Representative C2ST analyses commonly consider equal sample sizes \citep{lopez2017revisiting,rosenblatt2021better,gerber2023minimax}, or at least samples of the same asymptotic order \citep{kim2021classification,hediger2022use,cheng2022classification}.
Recent work has begun to consider two-sample testing under sample-size asymmetry \citep{chen2013ensemble,tian2026lottery}, but this regime remains largely underexplored for classifier-based testing procedures. 

This paper studies C2ST for complex, possibly non-Euclidean, data under severe sample imbalance. 
Our key idea is that, the surplus majority observations should not be viewed merely as a nuisance that complicates classifier training; rather, they can be exploited systematically as a statistical resource that can provide repeated testing evidence. 
Concretely, we partition the highly imbalanced dataset into multiple approximately balanced sub-datasets, each of which admits a more reliable classifier fit. 
We then evaluate these classifiers on a common held-out sample and aggregate their evidence while accounting for the dependence induced by shared observations.
Unlike balanced-ensemble approaches in imbalanced classification, which use multiple balanced learners primarily to improve predictive performance, our construction uses the resulting balanced fits as building blocks of a hypothesis-testing procedure and explicitly calibrates and aggregates dependent testing evidence from those balanced fits.
This strategy also fundamentally differs from standard imbalance remedies such as class weighting or oversampling: instead of modifying a single imbalanced classification problem, we repeatedly rebalance the data at the level of the testing procedure and then aggregate the resulting evidence across the subproblems. This rebalancing improves the reliability and power of C2ST while preserving its flexibility as to classifier choice and data modality. 

Building on this insight, we propose the \emph{Imbalance-Aware Deep Classifier Two-Sample Test} and its enhanced version, collectively referred to as \emph{(E)IAD-C2ST}. For computationally expensive deep classifiers, we further develop a scalable implementation strategy, using representation amortization to reduce the computational burden of classifier training. 
The design logic, and therefore the attribution of power gains, consists of three components: balanced classifier fitting, which prevents class imbalance from dominating each classifier fit; blockwise evidence aggregation, which governs the use of the majority surplus; and train-test symmetrization, which governs the use of the two train-test orientations.
The main contributions are summarized as follows.

\begin{enumerate}[leftmargin=*]

    \item \textbf{A unified, imbalance-aware C2ST framework for complex data.}

    EIAD-C2ST offers a unified extension of C2ST to the severely imbalanced regime for complex data while retaining all the practical appeal of C2ST.  When the two samples are balanced, IAD-C2ST reduces to the vanilla C2ST, while EIAD-C2ST further mitigates split-induced instability. 
    When one sample is substantially larger than the other, the majority sample is partitioned into blocks of comparable size to the minority sample, converting the original imbalanced testing problem into multiple approximately balanced blockwise C2ST problems. This avoids the interpolation of observations, making it particularly useful for non-Euclidean data without a readily available synthetic-interpolation mechanism. This construction treats the surplus majority observations not as a nuisance, but as a statistical resource for repeated evidence generation. 
    EIAD-C2ST retains the flexibility and broad applicability of C2ST. 
    The EIAD-C2ST framework is classifier-agnostic and can be paired with any backbone classifier suitable for the data modality, for example, classical logistic classifiers or multilayer perceptrons (MLPs) for tabular data, convolutional neural nets (CNNs) for images, graph neural networks (GNNs) for graph-structured data, recurrent neural networks (RNNs) for sequential data, and Transformers for text data.  

    \item \textbf{Rigorous theoretical guarantees on validity.} 

    We establish rigorous validity guarantees for the proposed (E)IAD-C2ST, importantly, without requiring the balanced-sample condition that standard C2ST procedures typically rely on.  
    This ensures the robust and reliable performance of (E)IAD-C2ST despite the presence of severe sample imbalance. A central challenge is that the blockwise tests share both the minority training observations and the held-out evaluation sample. We prove that the blockwise statistics and $p$-values generated by the imbalance-aware splitting scheme are exchangeable under the null hypothesis, despite this dependence. This exchangeability justifies valid $p$-value aggregation through exchangeable merging rules, despite the dependence induced by reusing the minority sample across blockwise tests. EIAD-C2ST further improves stability by repeating this construction across complementary train-test splits and combining the resulting evidence. 

    \item \textbf{Inheritance of consistent power and empirical power gains.}

    We also provide theoretical guarantees on power performance and characterize the conditions under which (E)IAD-C2ST achieves consistent power. Theorem~\ref{thm:power-IAD-C2ST} shows that IAD-C2ST inherits consistent power from any split C2ST whenever the evidence is strong enough, and Theorem~\ref{thm:size-power-EIAD-C2ST} further shows that EIAD-C2ST inherits consistent power from IAD-C2ST. Empirically, 
    simulations on Euclidean and graph data and a real-data application show that (E)IAD-C2ST delivers substantial empirical gains and typically achieves higher power than competing benchmark methods in severely imbalanced regimes, while maintaining Type-I error control. These results demonstrate that the proposed aggregation steps make more effective use of the available majority observations, preserving asymptotic consistency and substantially improving the practical performance of C2ST. 
    
\end{enumerate}

The rest of the paper is organized as follows.
Section~\ref{sec: method} studies the impact of sample imbalance on C2ST and introduces (E)IAD-C2ST together with its representation-amortized implementation.
Section~\ref{sec: theory} establishes the theoretical properties of the proposed procedures.
Section~\ref{sec: sim} presents simulation studies in Euclidean and non-Euclidean settings.
Section~\ref{sec: Real Dataset} reports an application to Bitcoin transaction data.
Section~\ref{sec:discussion} concludes with a discussion.
Proofs and additional details of the empirical studies are provided in the Appendices.

\section{Methods}\label{sec: method}

Let $(\Omega, d)$ denote a metric space with a metric $d$. Suppose $\{X_i\}_{i=1}^{n}$ is a sample of random objects taking values in $\Omega$, generated according to a Borel probability measure $\calP_{\calX}$ on $\Omega$, and that $\{Y_j\}_{j=1}^{m}$ is another sample of $\Omega$-valued random objects generated analogously according to a Borel probability measure $\calP_{\calY}$. We study the problem of two-sample testing on
\[
H_0:\calP_{\calX}=\calP_{\calY},
\qquad
H_1:\calP_{\calX}\neq \calP_{\calY}.
\]
in metric space $(\Omega, d)$ when the samples are complex and possibly non-Euclidean.

We begin by assigning a dichotomous label ($\{\pm 1\}$) to the observed samples from the two groups, which gives $\calD_{X} := \{(X_i, 1)\}_{i=1}^{n}$ (positive class) and $\calD_{Y} := \{(Y_j, -1)\}_{j=1}^{m}$ (negative class). Without loss of generality, we assume $n \leq m$. Here, we allow the samples from the two groups to be highly imbalanced. Let $f: \Omega \rightarrow \{-1, 1\}$ denote a classifier that maps a datapoint $Z\in \Omega$ to a label taking values in $\{-1, 1\}$. Here, \textbf{$f$ could be any classifier suitable for the data type} in $\Omega$.
Intuitively, if the null hypothesis $H_0:\calP_{\calX}=\calP_{\calY}$ is true,  any classifier trying to distinguish between two samples would behave like a random guess. In other words, a significantly better-than-chance performance of a classifier provides evidence that the two samples should be statistically different. 
To assess the out-of-sample performance of a classifier, let
$g:\Omega\to\{-1,1\}$ be a measurable classifier, which may be
data-dependent. Then we define the averaged class-wise population classification error as
\begin{equation}
\mathcal E_g
:=
\frac{\mathcal E_{1,g}+\mathcal E_{-1,g}}{2}.
\label{eq:true-classwise-error}
\end{equation}
where 
\[
\mathcal E_{1,g}
:=
\int_{\Omega}
\mathbf 1\{g(z)=-1\}\,d\mathcal P_X(z),
\qquad
\mathcal E_{-1,g}
:=
\int_{\Omega}
\mathbf 1\{g(z)=1\}\,d\mathcal P_Y(z),
\] 
When $g$ is data-dependent, these quantities are random through the
realized fitted classifier $g$; the integrals are evaluated while holding
that realized classifier fixed. When $g$ is deterministic, they reduce
to the usual class-wise misclassification probabilities.

Here, the averaged class-wise classification error provides a symmetric measure of classification performance that does not favor either class. This is particularly important when samples are imbalanced, where the pooled misclassification rate can be dominated by the larger sample class and may therefore provide a misleading summary of discriminating ability.

To estimate $\calE_f$, we adopt a commonly used strategy by randomly dividing the observed samples into training and testing sets. We randomly partition $\calD_{X}$ into 
$\calD_{X}^{tr}$ and $\calD_{X}^{te}$, and similarly $\calD_{Y}$ into $\calD_{Y}^{tr}$ and $\calD_{Y}^{te}$, and define $\calD^{tr}:= \{\calD_{X}^{tr}, \calD_{Y}^{tr}\}$ for training and $\calD^{te}:= \{\calD_{X}^{te}, \calD_{Y}^{te}\}$ for testing.  
We use $n_1=|\calD_X^{tr}|$, $m_1=|\calD_Y^{tr}|$,  $n_2=|\calD_X^{te}|$, and $m_2=|\calD_Y^{te}|$,
to denote the sample sizes of $\calD_{X}^{tr}$, $\calD_{Y}^{tr}$, $\calD_{X}^{te}$ and $\calD_{Y}^{te}$, respectively. 

Let $\texttt{Train}(\cdot)$ denote the training routine for a chosen backbone classifier $f$, and let $\widehat{f}= \texttt{Train}(\calD^{tr})$ denote the classifier trained on $\calD^{tr}$. We assume that the learned classifier $\widehat{f}$ is sufficiently well-trained (formalized in Assumptions \ref{assump:null-nondegeneracy} and \ref{assump:learning-algorithm}), in the sense that the class-wise classification errors based on $\widehat{f}$ are asymptotically accurate for approximating the true class-wise classification errors.  

An estimate of the averaged class-wise classification error is obtained using $\calD^{te}$ as 
\begin{align}
\widehat{\calE}_{\widehat{f}} := (\widehat{\calE}_{1, \widehat{f}}+\widehat{\calE}_{-1, \widehat{f}})/2,
\label{auc_f}
\end{align}
where 
$$
\widehat{\calE}_{1, \widehat{f}}:= n_2^{-1} \sum\limits_{(z, l) \in \calD_{X}^{te} } \mathbf{1}\{ \widehat{f}(z) = -1 \}  \quad 
\text{ and } \quad
\widehat{\calE}_{-1, \widehat{f}}:= m_2^{-1} \sum\limits_{(z, l) \in \calD_{Y}^{te} } \mathbf{1}\{ \widehat{f}(z) = 1 \}.
$$
Thus, $\widehat{\calE}_{1,\widehat f}$ estimates the proportion of positive-class test observations that are incorrectly predicted as negative, while $\widehat{\calE}_{-1,\widehat f}$ estimates the proportion of negative-class test observations that are incorrectly predicted as positive. Under $H_0$, where the two distributions coincide, one expects $\widehat{\calE}_{\widehat f}$ to be close to $0.5$; under $H_1$, an informative classifier should yield $\widehat{\calE}_{\widehat f}<0.5$. A standardized statistic for the Classifier Two-Sample Test (C2ST) \citep{lopez2017revisiting} can then be naturally constructed as follows. 
\begin{equation}
\label{stat:standardized-classification-error}
    T_{\widehat{f}} = \frac{\widehat{\calE}_{\widehat{f}} - 0.5}{\sqrt{ \frac{1}{4n_2} \widehat{\calE}_{1, \widehat{f}} (1-\widehat{\calE}_{1, \widehat{f}}) + \frac{1}{4m_2} \widehat{\calE}_{-1, \widehat{f}} (1-\widehat{\calE}_{-1, \widehat{f}})  }}
\end{equation}
When the estimated variance in the denominator of \eqref{stat:standardized-classification-error} is zero, we define the corresponding \(p\)-value according to the direction of the numerator: \(p=0\) if the numerator is strictly negative and \(p=1\) otherwise. This convention preserves the one-sided direction of the test while assigning no evidence against the null in the indeterminate \(0/0\) case. 

Note that since evidence against $H_0$ corresponds to classification error substantially below the chance level, smaller values of $T_{\widehat f}$ provide stronger evidence against the null. 
Existing studies on theoretical properties of the C2ST are established under conditions: 
\begin{itemize}[label={}]
    \item[\textbf{(C1)}]\condlabel{cond:sample-size-ratio} a sample-size-ratio condition that regulates the ratio of sample sizes: there exists a constant $c\in(0,1)$ such that $n/(n+m)\rightarrow c$, ruling out highly imbalanced cases;
    \item[\textbf{(C2)}]\condlabel{cond:sample-splitting-ratio} a sample-splitting-ratio condition that regulates the balance between training and testing sets: there exists a constant $\kappa \in (0,1)$, such that $(n_1+m_1)/(n+m) \rightarrow \kappa$.
\end{itemize}

With Conditions (\ref{cond:sample-size-ratio})--(\ref{cond:sample-splitting-ratio}) and some regularity assumptions on the classifier, \cite{kim2021classification} showed $ T_{\widehat{f}} \overset{d}{\rightarrow} N(0,1)$ as $n, m \rightarrow \infty$.
Then, the $p$-value of $T_{\widehat{f}}$ for testing $H_0$ can be computed using 
\begin{equation} \label{pval:T-balanced}
    pval(T_{\widehat{f}}) = \Phi(T_{\widehat{f}})
\end{equation}
where $\Phi(\cdot)$ is the cdf of standard normal distribution. The test rejects $H_0$ at the significance level $\alpha$ if $T_{\widehat{f}} < -z_{1-\alpha}$ where $z_{1-\alpha}=\Phi^{-1}(1-\alpha)$. %

\subsection{The Impact of Sample Imbalance} \label{subsec:impact}

We first study the impact of sample imbalance.
A classifier trained on $\mathcal{D}^{tr} = (\mathcal{D}_X^{tr}, \mathcal{D}_Y^{tr})$ optimizes an empirical classification risk:
$$
 \frac{1}{n_1 + m_1} \left[ \sum_{i=1}^{n_1} \ell(f(X_i), 1) + \sum_{j=1}^{m_1} \ell(f(Y_j), -1) \right]  
$$
where $\ell(\cdot,\cdot)$ is a classification loss function, e.g., the cross-entropy loss.
When $n_1 \ll m_1$, this objective is heavily imbalanced and is dominated by the majority class. As a result, the classifier can achieve a small training loss by simply favoring the majority class, which may reduce its sensitivity to distributional differences. 
Consequently, the trained classifier $\widehat{f}$ may be poorly calibrated, thereby weakening the power of the resulting test. 

\begin{figure}[htb]
    \centering
    \includegraphics[width=0.9\linewidth,height=2in]{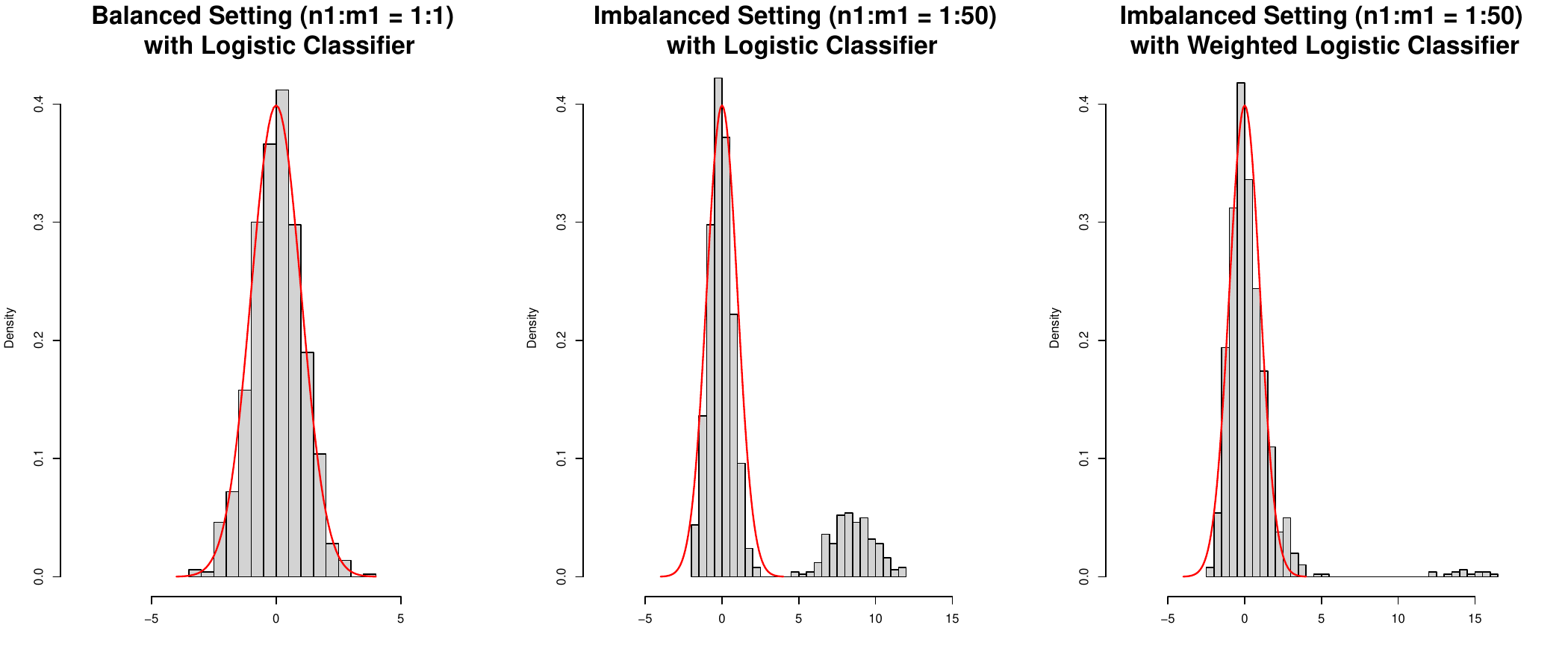}
    \vspace{-2ex}
    \caption{Histograms of the statistic \eqref{stat:standardized-classification-error} in a sample-balanced setting (left) and imbalanced settings (middle and right) under $H_0$ from 1,000 replications. Red lines superimposed in the plots are the probability density of $N(0,1)$. 
    The empirical distribution aligns with the theoretical asymptotic distribution in the balanced setting, whereas there are apparent misalignments in the imbalanced setting. }
    \label{fig:plot_asymp_dist}
\end{figure}

Figure \ref{fig:plot_asymp_dist} plots the empirical distribution of test statistic \eqref{stat:standardized-classification-error} under simulated scenarios with different sample-size ratios. 
As shown in the plot, the empirical distribution aligns with the theoretical asymptotic distribution in the balanced setting, whereas there is an apparent misalignment in the imbalanced setting.
Cost-sensitive classification methods, such as weighted logistic regression, may improve empirical performance; however, they do not directly address the fundamental challenge of sample imbalance and may still produce apparent ``outliers'' that deviate from the standard normal distribution.
Such shifts in distribution lead to deteriorated performance when sample sizes are highly imbalanced, showing an overly conservative size under the null and loss of power under the alternative.
This motivates the development of an improved method that explicitly accounts for sample-size imbalance. 

\vspace{2ex}

\subsection{An Enhanced Imbalance-Aware Deep Classifier Two-Sample Test (EIAD-C2ST)}
\label{sec: EIAD-C2ST}

\begin{figure}[p]
    \centering
    \includegraphics[width=0.9\linewidth]{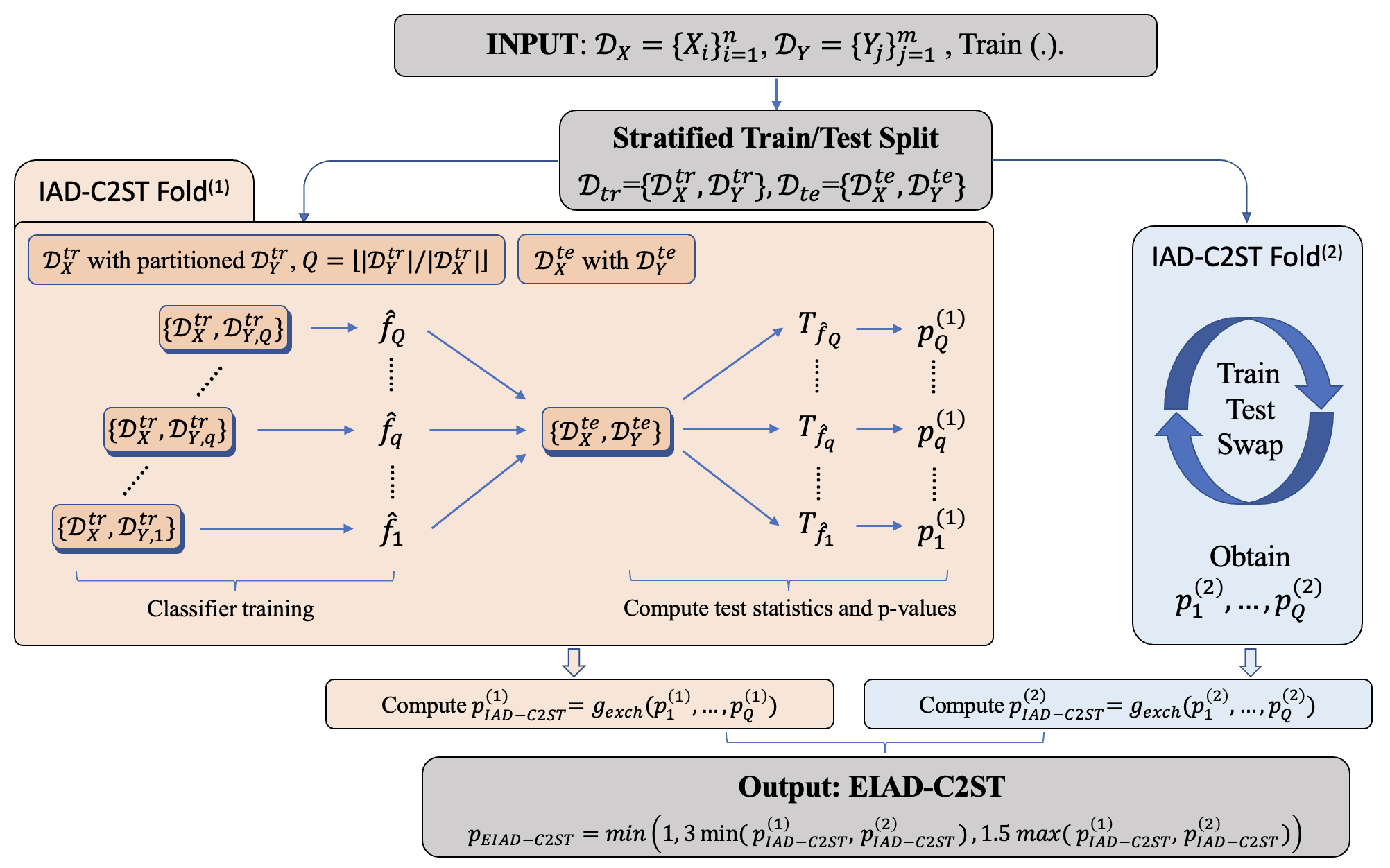}
    \caption{\small
    Flowchart of EIAD-C2ST. The procedure first performs a stratified random split, then constructs two split-specific IAD-C2ST statistics by partitioning the majority sample into approximately balanced slices, training and evaluating classifiers on the resulting blockwise subproblems, and merging the corresponding blockwise $p$-values. The final EIAD-C2ST $p$-value is obtained by aggregating the two IAD-C2ST $p$-values as in \eqref{eq:pval-EIAD-C2ST}.}
    \label{fig: eiad}

    \begin{minipage}{\linewidth}
    \begin{algorithm}[H]
\caption{EIAD-C2ST}
\label{alg:eiaD-C2ST}
\begin{algorithmic}[1]
\STATE \textbf{Input:} Samples $\mathcal{D}_X=\{X_i\}_{i=1}^{n}$, $\mathcal{D}_Y=\{Y_j\}_{j=1}^{m}$, a training routine $\texttt{Train}(\cdot)$  
\STATE \textbf{Output:} the $p$-value of EIAD-C2ST 

\STATE Perform a stratified random split which partitions $\mathcal{D}_X$ into two halves $\mathcal{D}_X^{tr}$ and $\mathcal{D}_X^{te}$, and $\mathcal{D}_Y$ into two halves $\mathcal{D}_Y^{tr}$ and $\mathcal{D}_Y^{te}$. Denote $\calD^{tr} = \{\mathcal{D}_X^{tr}, \mathcal{D}_Y^{tr}\}$ and $\calD^{te} = \{\mathcal{D}_X^{te}, \mathcal{D}_Y^{te}\}$. 

\STATE Use $\calD^{tr}$ to train classifiers and $\calD^{te}$ to construct test statistics. Set $Q=\left\lfloor |\mathcal{D}_Y^{tr}|/|\mathcal{D}_X^{tr}|\right\rfloor$. 
Randomly partition $\mathcal{D}_Y^{tr}$ 
into $Q$ disjoint splits $\{\mathcal{D}_{Y,q}^{tr}\}_{q=1}^{Q}$ 
with approximately equal size.
\FOR{$q=1,\ldots,Q$}
    \STATE Construct $\mathcal{D}_{q}^{tr,(1)}:=\{\mathcal{D}_X^{tr},\mathcal{D}_{Y,q}^{tr}\}$ and train $\widehat f_q^{(1)}\leftarrow\texttt{Train}(\mathcal{D}_{q}^{tr,(1)})$.
    \STATE Evaluate $\widehat f_q^{(1)}$ on $\mathcal D^{te}=\{\mathcal D_X^{te},\mathcal D_{Y}^{te}\}$, compute $T_q^{(1)}$ by \eqref{stat:standardized-classification-error}, and set $p_q^{(1)}\leftarrow \Phi(T_q^{(1)})$.
\ENDFOR
\STATE Compute
$
p_{\rm IAD-C2ST}^{(1)} = g_{\rm exch} (p_1^{(1)},\ldots,p_{Q}^{(1)} )
$
with $g_{\rm exch}(\cdot)$ defined by \eqref{pval:HMP-exch-rule}. 

\STATE Swap the role of $\calD^{tr}$ and $\calD^{te}$, repeat Lines 4-8 to obtain $p_1^{(2)},\ldots,p_{Q}^{(2)}$, and compute
$
p_{\rm IAD-C2ST}^{(2)} = g_{\rm exch} (p_1^{(2)},\ldots,p_{Q}^{(2)} ).
$

\STATE Compute $p_{\rm EIAD-C2ST}$ by \eqref{eq:pval-EIAD-C2ST}.
\STATE \textbf{Return:} $p_{\rm EIAD-C2ST}$  
\end{algorithmic}
\end{algorithm}

    \end{minipage}
\end{figure}

Hereafter, we propose a new imbalance-aware testing procedure that tackles the challenge of sample imbalance in C2ST. 
The proposed procedure is specifically designed as a two-step framework to address two distinct sources of power loss and instability.
The first step addresses the fundamental challenge of class imbalance by converting a single severely imbalanced classification problem into a collection of approximately balanced subproblems, each admitting a more reliable classifier fit.
The second step reduces the split-induced instability inherent to the train-test assignment 
by repeating the construction with the roles of training and testing sets interchanged, and then aggregating the resulting evidence using an efficient combination rule. 
The complete algorithm of EIAD-C2ST is summarized in Algorithm \ref{alg:eiaD-C2ST}, with a schematic overview provided in Figure~\ref{fig: eiad}.

Without loss of generality, we assume $n \leq m$ (i.e., $\calD_{X}$ is the minority class and $\calD_{Y}$ is the majority class).
To ensure a representative sampling of classes, we consider a stratified splitting strategy that preserves the sample size ratio of the two groups. 
For simplicity, we randomly split $\calD_{X}$ into two equal halves $\calD_{X}^{tr}$ and $\calD_{X}^{te}$, $\calD_{Y}$ into two equal halves $\calD_{Y}^{tr}$ and $\calD_{Y}^{te}$.

\paragraph{Step 1. (Handling sample imbalance.) }

Recall that the training data is $\calD^{tr}:= \{\calD_{X}^{tr}, \calD_{Y}^{tr}\}$ with sample sizes $n_1$ and $m_1$ of each group.
Let $Q := \lfloor m_1/n_1 \rfloor \geq 1$. Here, we consider $Q$ as a deterministic number.
We randomly partition $\calD_{Y}^{tr}$ into $Q$ splits of approximately equal size, denoted by $\calD_{Y, q}^{tr}$ with its sample size being $m_{1,q}$, $q=1,\ldots, Q$.
Then for each $q$, we train a classifier, denoted by $\widehat{f}_q = \texttt{Train}(\calD^{tr}_{q})$, based on $\calD^{tr}_{q}:= \{\calD_{X}^{tr}, \calD_{Y, q}^{tr}\}$, and compute the test statistic $T_q := T_{\widehat{f}_q}$ associated with $\widehat{f}_q$ using the testing set $\calD^{te} = \{\calD_{X}^{te}, \calD_{Y}^{te}\}$. 

By construction, we have $n_{1}\approx m_{1,q}$ for all $q\in \{1,\dots,Q\}$.
Therefore, balanced blocks address classifier fitting, and the unequal-sample Studentized analysis addresses calibration on the held-out data.
The $p$-value associated with $T_{q}$ can be obtained via 
$$p_q := pval(T_{q}) = \Phi(T_{q}).$$ 

To aggregate information from the $Q$ $p$-values efficiently, we investigate their dependence. It is noted that the training of $\widehat{f}_q$ uses the same $\calD_X^{tr}$ but different $\calD_{Y,q}^{tr}$ for $q = 1, \ldots, Q$, and moreover, the construction of $T_q$ uses the same testing set $\calD^{te} = \{\calD_{X}^{te}, \calD_{Y}^{te}\}$.
Hence, the $p$-values resulting from the $Q$ tests are inherently dependent. In Theorem~\ref{thm:exchangeablity}, we show that this dependence has a specific, tractable structure: the vector $(p_1,\ldots,p_Q)$ is
exchangeable, which enables efficient aggregation through exchangeable combination rules \citep{liu2022multiple,choi2023averaging,gasparin2025combining}. 

Let $g_{exch}(\cdot)$ denote a $p$-value combination rule that can accommodate the exchangeability among these $p$-values.
We propose an \textbf{Imbalance-Aware Deep Classifier Two-Sample Test (IAD-C2ST)} that aggregates information via  
\begin{align} 
\label{pval:combined}
p_{\rm IAD-C2ST} = 
     & g_{\rm exch}(p_1, \ldots, p_q, \ldots, p_Q),
\end{align}
in which
\begin{align}\label{pval:HMP-exch-rule}
    g_{\rm exch}(p_1, \ldots, p_q, \ldots, p_Q) = \left\{
    \begin{aligned}
        & \quad p_1, \quad  & \text{if } Q=1; \\ 
        & \quad \min_{l=1,\ldots, Q} \left\{ \min_{s=1,\ldots,l} \left( \frac{l\cdot C_Q}{s} + 1 \right) \cdot H_{l,s} \right\}, \quad  & \text{if } Q\geq 2.
    \end{aligned}
    \right.
\end{align}
with $C_Q = \log Q+ \log\log Q + 1$ and $H_{l,s}$ being the partial harmonic mean of the $s$ smallest $p$-values among $\{p_1, \ldots, p_l\}$. 
When samples are balanced with $Q=1$, the IAD-C2ST coincides with the vanilla C2ST \citep{lopez2017revisiting}. 
The IAD-C2ST rejects $H_0$ at the significance level $\alpha$ if $$p_{\rm IAD-C2ST} \leq \alpha.$$

\paragraph{Step 2. (Reducing the split-induced instability.) }
Data-splitting, despite its popularity, is often criticized for using only part of the sample at each stage, which may reduce efficiency relative to procedures that reuse the full sample more directly \citep{meinshausen2009p,fan2012variance,romano2019multiple,liu2022multiple}. 
In addition to this loss of efficiency, a split-based classifier test may also be sensitive to the arbitrary assignment of the two split halves to the training and testing roles. 
This sensitivity can be particularly acute when the minority class is small, because the realized composition of each half can have a noticeable effect on both classifier fitting and test-set evaluation.

We take additional steps to reduce this artificial instability induced by the train-test assignment.
To this end, we symmetrize the procedure by swapping the roles of the training and testing sets, yielding two orientation-specific $p$-values, denoted by $p_{\rm IAD-C2ST}^{(1)}$ and $p_{\rm IAD-C2ST}^{(2)}$. 
Under the uniform stratified equal-half split and the symmetric
randomization construction, these two $p$-values are exchangeable; see Proposition~\ref{prop:orientation-exchangeability-EIAD-C2ST}.
This finite-sample property holds under both the null and
alternative hypotheses and formalizes the fact that neither
orientation is statistically privileged, although their realized
$p$-values can differ substantially. We therefore combine the
two results using a rule that is invariant to their labels.
Specifically, we construct the \textbf{Enhanced Imbalance-Aware Deep Classifier Two-Sample Test
(EIAD-C2ST)}: 
\begin{align} 
     p_{\rm EIAD-C2ST} := & \ p_{\rm EIAD-C2ST}(w) \label{eq:pval-EIAD-C2ST} \\  \nonumber
     = & \min\left\{ 1, \frac{\min(p_{\rm IAD-C2ST}^{(1)}, p_{\rm IAD-C2ST}^{(2)})}{w}, \frac{\max(p_{\rm IAD-C2ST}^{(1)}, p_{\rm IAD-C2ST}^{(2)})}{1-w} \right\}
\end{align}
where $0 < w \leq 0.5$ is a prespecified evidence-balance parameter.
The EIAD-C2ST rejects $H_0$ at the significance level $\alpha$ if $$p_{\rm EIAD-C2ST}(w)  \leq \alpha.$$

EIAD-C2ST can be viewed as a validity-preserving symmetrization of IAD-C2ST with respect to the train-test assignment. Rather than relying on a single arbitrary orientation of a stratified split, it aggregates evidence from the two equally legitimate orientations. This construction reduces sensitivity to the train-test assignment while preserving asymptotic type-I error control. Under alternatives, EIAD-C2ST is consistent whenever at least one of the two assignment-specific IAD-C2ST procedures is consistent. These theoretical guarantees are established in Theorem~\ref{thm:size-power-EIAD-C2ST}. In finite samples, the aggregation can also mitigate power loss when one orientation of the split is substantially less favorable than the other.

\paragraph{Exchangeability and orientation invariance.}

The proposed symmetrization of EIAD-C2ST is supported by two properties that formalize why the two orientations should be treated on equal footing.
First, as presented formally in  Proposition~\ref{prop:orientation-exchangeability-EIAD-C2ST}, 
the two orientation-specific $p$-values are exchangeable.
This is a distributional property: neither orientation is
statistically privileged, although the two realized $p$-values
can differ substantially.
Second, let $G_w(u,v)$ denote the combination map on the
right-hand side of \eqref{eq:pval-EIAD-C2ST}. This map satisfies
the deterministic symmetry: 
$
    G_w(u,v)=G_w(v,u)$ for $
    (u,v)\in[0,1]^2.
$
Consequently, once the two orientation-specific fits have been
computed, relabeling their orientations leaves the combined
$p$-value and rejection decision unchanged. This invariance
holds for every pair of inputs and does not require
exchangeability.
Thus, exchangeability establishes the statistical equivalence
of the two orientation labels, whereas symmetry of the combination rule makes the reported result invariant to those labels. The
construction removes the need to choose between the two
orientations of a given partition, without eliminating
variability arising from the partition itself or from
classifier fitting.

 \paragraph{Interpretation of the parameter $w$.} Let $p_{\min} := \min(p_{\rm IAD-C2ST}^{(1)}, p_{\rm IAD-C2ST}^{(2)})$ and $p_{\max} := \max(p_{\rm IAD-C2ST}^{(1)}, p_{\rm IAD-C2ST}^{(2)})$ denote the ordered $p$-values.
The parameter $w$ allocates the rejection threshold between two forms of evidence: strong evidence from either train-test orientation, represented by $p_{\min} \leq w\alpha$, and corroborating evidence from both orientations, represented by $p_{\max} \leq (1-w)\alpha$. Every prespecified $w \in (0,0.5]$ preserves the asymptotic validity of $p_{\rm EIAD-C2ST}(w)$. 
To choose $w$, we compare EIAD-C2ST with two natural benchmark
$p$-values $2p_{\min}$ and $p_{\max}$, corresponding to isolated-evidence from either orientation and corroborating-evidence from both orientations, respectively.
We select $w$ to minimize the worst-case multiplicative inflation relative to the more favorable of these two benchmarks:
\begin{equation*} 
\inf_{0<w\leq 1/2}
\sup_{0<p_{(1)}\leq p_{(2)}\leq 1}
\frac{p_{\rm EIAD-C2ST}(w)}
     {\min\{2p_{min},p_{max}\}}
=
\inf_{0<w\leq 1/2}
\max\left\{
\frac{1}{2w},
\frac{1}{1-w}
\right\}
=
\frac{3}{2}.
\end{equation*}
The minimum is uniquely attained at $w=1/3$, for which the two
multiplicative penalties are balanced.
This is precisely the two-input Hommel combination rule \citep{hommel1983tests,vovk2022admissible}. 
We use $w=1/3$ throughout.

\subsection{Representation-Amortized Implementation of (E)IAD-C2ST: A Scalable Variant using Probing Classifiers}\label{sec:scalable-EIAD-C2ST}

The proposed (E)IAD-C2ST procedures train slice-specific classifiers on approximately balanced subproblems.
Each balanced slice in EIAD-C2ST retrains a \emph{complete} classifier from scratch, 
and the symmetrization step repeats this for the swapped train--test assignment.
The number of slices $Q$ grows with the imbalance ratio, so producing a single EIAD-C2ST $p$-value requires $2Q$ classifier fits in total.  
For simple classifiers like logistic regression, the training process is straightforward and these repeated fits are computationally inexpensive. 
When the backbone is a deep neural network, such as a GNN, these repeated fits dominate the overall cost and make the procedure expensive to deploy at scale.

We introduce a scalable variant for EIAD-C2ST using \emph{representation amortization} for computationally expensive deep classifiers, which preserves the testing logic of EIAD-C2ST while amortizing the cost of per-slice deep retraining through cheap probing. The key idea is to factor the classifier into a representation map and a lightweight head, learning the nonlinear representation only once within each train-test orientation and then fitting lightweight slice-specific probing classifiers on top of the frozen representation. 
For a deep classifier $f_q$, we first decompose the function into $f_q=h_q\circ\phi$,  
where $\phi$ is the representation map learned by the hidden layers and $h_q$ is the final task-specific prediction head.  
This decomposition mirrors the representation-head architectural view of deep neural networks employed in many modern deep learning paradigms, where pretrained or learned representations are often frozen and used as feature extractors for downstream linear probing or transfer learning tasks \citep{donahue2014decaf,yosinski2014how,alain2017understanding,belinkov-2022-probing}.
In our setting, we first learn the representation $\phi$ once using the full training sample and hold it fixed; only the prediction head $h_q$ is trained using the sliced subsample. This decoupling thereby provides a computationally efficient variant of the full slice-specific deep network training. 

\begin{figure}[!tbp]
    \centering
    \includegraphics[width=0.9\linewidth]{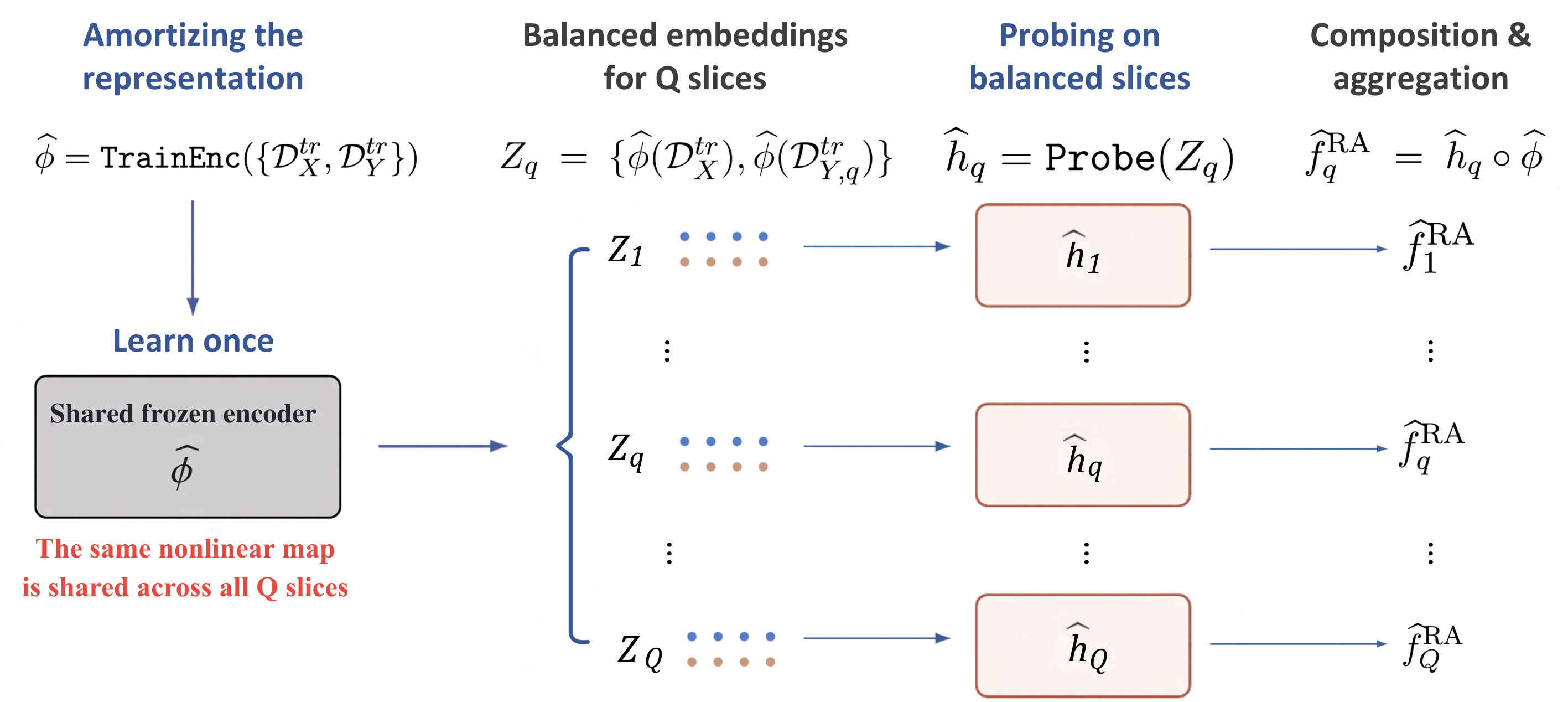}
    \vspace{-2ex}
    \caption{Representation-amortized implementation of EIAD-C2ST within one train--test orientation. A single nonlinear encoder $\widehat{\phi}$ is learned from the full training sample and then frozen. For each balanced slice $q$, a lightweight probe $\widehat{h}_q$ is fitted on the frozen embeddings $Z_q=\{\widehat{\phi}(\calD_X^{tr}),\widehat{\phi}(\calD_{Y,q}^{tr})\}$, yielding $\widehat{f}^{\text{RA}}_{q}=\widehat{h}_q\circ\widehat{\phi}$. The same encoder is reused across all $Q$ slices; the swapped orientation and $p$-value aggregation follow the procedure in Algorithm \ref{alg:eiaD-C2ST}.}
    \label{fig:ra-eiad-representation}
\end{figure}

Specifically, conditional on the realized split, we implement the EIAD-C2ST with representation amortization as follows.
\begin{enumerate}[leftmargin = *]
\item[(i)] \textbf{(Amortizing the representation.)} Train the deep backbone classifier \emph{once} on the full training data $\calD^{tr}=\{\calD_X^{tr},\calD_Y^{tr}\}$, and retain its representation map as a \emph{frozen encoder}, which is held fixed throughout the remaining steps. Specifically,
\begin{equation*}
    \widehat{\phi}:= \texttt{TrainEnc}(\{\calD_X^{tr},\calD_Y^{tr}\}).
\end{equation*}

\item[(ii)] \textbf{(Probing on balanced slices.)} Partition $\calD_Y^{tr}$ into $Q$ slices $\{\calD_{Y,q}^{tr}\}_{q=1}^{Q}$ as in Section~\ref{sec: EIAD-C2ST}. For each $q$, train a linear probing classifier on the frozen embeddings of the balanced block
$\{\widehat{\phi}(\calD_X^{tr}),\widehat{\phi}(\calD_{Y,q}^{tr})\}$, denoted by 
\begin{equation*}
    \widehat{h}_q := \texttt{Probe}(\{\widehat{\phi}(\calD_X^{tr}),\widehat{\phi}(\calD_{Y,q}^{tr})\}).
\end{equation*}

\item[(iii)] \textbf{(Composition and aggregation.)} The slice-specific classifier is then obtained by 
\begin{equation*}
  \widehat{f}^{\text{RA}}_{q} \;=\; \widehat{h}_q\circ\widehat{\phi}.  
\end{equation*}

The test statistic $T^{\text{RA}}_{q}$ and $p$-value $p^{\text{RA}}_{q}$ are constructed following the same procedure as in \eqref{stat:standardized-classification-error} and \eqref{pval:T-balanced}, 
and then aggregated via the same exchangeable rule. 
\end{enumerate}

This construction replaces Lines 5--8 of Algorithm~\ref{alg:eiaD-C2ST}. 
The representation-amortized implementation provides a practical acceleration that avoids retraining a deep network for every data slice. In Theorem \ref{thm:size-power-RA-EIAD-C2ST}, we show that (E)IAD-C2ST with representation amortization shares the same theoretical properties as (E)IAD-C2ST with full fitting. Figure~\ref{fig:ra-eiad-representation} illustrates the shared encoder and slice-specific probes within one train--test orientation.

\section{Theoretical Properties}\label{sec: theory}

In this section, we establish theoretical guarantees for IAD-C2ST and EIAD-C2ST. 
Throughout this section, the sample sizes $(n,m)$ index a sequence of
two-sample testing problems satisfying
$\min\{n,m\}\to\infty.$
For the problem indexed by $(n,m)$, define
$Q_{n,m}:= \left\lfloor{m_1}/{n_1}\right\rfloor$
and
$
n_{\min}
:=
\min\{n_2,m_2\}.
$
Accordingly,
the blockwise statistics and $p$-values form 
$\{T_{n,m,q}:q\in[Q_{n,m}]\}$
and
$\{p_{n,m,q}:q\in[Q_{n,m}]\}.$
For notational simplicity, we suppress the indices $(n,m)$ whenever no
confusion can arise.
We allow $Q_{n,m}$ to diverge as $\min\{n,m\}\to\infty$. 
When $Q_{n,m}=1$, there is no majority-sample surplus to partition,
and IAD-C2ST coincides with the vanilla C2ST. Hence, throughout the theoretical analysis, we focus on the nontrivial
imbalance regime $Q_{n,m}\ge2$ for all sufficiently large $(n,m)$.

Theorem \ref{thm:exchangeablity} proves the exchangeability of $\{T_1, T_2, \ldots, T_{Q_{n,m}}\}$ and $\{p_1, p_2, \ldots, p_{Q_{n,m}}\}$, laying the theoretical foundations for the proposed methods. Theorems~\ref{thm:size-IAD-C2ST}--\ref{thm:size-power-EIAD-C2ST} establish theoretical guarantees on the asymptotic size and power of IAD-C2ST and EIAD-C2ST, importantly, without requiring the standard proportional-sample-size
condition $n/(n+m)\to c\in(0,1)$. In particular, the imbalance ratio may
diverge, subject to an explicit condition connecting the growth of $Q_{n,m}$
to the marginal calibration accuracy of the blockwise $p$-values. Theorems~\ref{thm:exchangeablity-ra}--\ref{thm:size-power-RA-EIAD-C2ST} then extend these guarantees to the representation-amortized implementation, justifying the theoretical validity of these more scalable alternatives.

Let $ \mathcal S =
(\mathcal D_X^{tr},\mathcal D_X^{te},
 \mathcal D_Y^{tr},\mathcal D_Y^{te}) $
denote a stratified train-test split of the observed samples. 
Conditional on the realized split $\mathcal S=s$, the training and testing sets $\mathcal D_X^{tr}(s), 
\mathcal D_Y^{tr}(s), 
\mathcal D_X^{te}(s),$
and
$\mathcal D_Y^{te}(s)$ are treated as fixed. To simplify notation, we hereafter suppress the dependence on $s$ and write $\mathcal D_X^{tr}, \mathcal D_X^{te}, 
\mathcal D_Y^{tr}, \mathcal D_Y^{te}$ in place of $\mathcal D_X^{tr}(s), 
\mathcal D_X^{te}(s),  \mathcal D_Y^{tr}(s),  \mathcal D_Y^{te}(s)$  with the conditioning on a realized split  $\calS = s$ throughout. 

\begin{definition}[Fold Assignment Mapping]
Conditional on the realized split \(\mathcal S=s\), define an assignment map
$$
A:\mathcal D_Y^{\mathrm{tr}} \to \{1,\ldots,Q\},
$$
which maps each observation to a fold label. For each observation
\(y\in \mathcal D_Y^{\mathrm{tr}}\), \(A(y)=q\) means that \(y\) is assigned
to fold \(q\).
Given an assignment map \(A\), the induced training blocks are
$
\mathcal D_{Y,q}^{\mathrm{tr}}(A)
=
\{y\in \mathcal D_Y^{\mathrm{tr}}: A(y)=q\}, 
q=1,\ldots,Q.
$
Let \(\Omega_A(s)\) denote the admissible assignment space, namely the
collection of all assignment maps
$
A:\mathcal D_Y^{\mathrm{tr}} \to \{1,\ldots,Q\}
$
such that the induced blocks
\(\{\mathcal D_{Y,q}^{\mathrm{tr}}(A)\}_{q=1}^Q\) satisfy the required
partition and balance constraints, and, if applicable, any additional admissibility constraints.
\end{definition}

\begin{assumpA}[Uniform and Label-Symmetric Assignment] \label{assump:uniform-assignment}
Conditional on the realized split \(\mathcal S=s\), the fold assignment map
\(A\) is drawn uniformly from the admissible assignment space
\(\Omega_A(s)\). 
Moreover, for every permutation $\pi$ of $\{1,\ldots,Q\}$, the map $a \mapsto \pi\circ a$ is a bijection from $\Omega_A(s)$ onto itself.
\end{assumpA}

Each classifier $\widehat f_q$ is obtained by applying the same training routine to $\mathcal D^{tr}_q$, namely
$ \widehat f_q = \texttt{Train}(\mathcal D^{tr}_q),  q\in\{1,\ldots,Q\}. $
Thus, the same learning algorithm is used across all folds. One additional
source of randomness, specifically, the algorithmic randomness, also warrants explicit quantification.

\begin{definition}[Algorithmic Randomness]
Many modern machine learning algorithms, e.g., deep neural networks, are inherently randomized: different realizations of the internal randomness may produce different fitted classifiers even when the training data are identical. 
Such randomness may arise from random initialization, stochastic minibatch
ordering, dropout, or other pseudo-random operations used during training.
To characterize this randomness, let
$R_q\in\mathcal R$ denote the collection of all random quantities generated
internally by the learning algorithm when fitting the classifier associated with
fold $q$. We refer to $R_q$ as the algorithmic randomness.
The precise content of $R_q$ depends on the learning algorithm; see
Remark~\ref{rmk:algorithmic-randomness} for further discussion.
\end{definition}

Formally, the fitted classifier can be represented as
    $$\widehat{f}_q = \texttt{Train}(\calD_q^{tr})  := \Xi(\calD^{tr}_X, \calD^{tr}_{Y,q}, R_q),$$
    and the resulting test statistic can be written as
    $$T_q := \Psi(\widehat{f}_q, \calD_X^{te}, \calD_Y^{te}).$$
    Equivalently, 
    \begin{equation}\label{eq:Tq-representation1}
        T_q := \calT(\calD^{tr}_X, \calD^{tr}_{Y,q}, \calD_X^{te}, \calD_Y^{te}, R_q)
    \end{equation}
where $\Xi$, $\Psi$, $\calT$ are measurable functions that do not depend on the fold label $q$.

\begin{remark}[The Presence/Absence of Algorithmic Randomness in  Randomized/Deterministic Learners]\label{rmk:algorithmic-randomness}
    Many modern machine learning algorithms are inherently randomized. 
    For example, $R_q$ may include random parameter initialization, stochastic gradient descent minibatch ordering, dropout masks, early-stopping randomness, or any other pseudo-random operations used during training. Such randomness is common in modern machine learning methods, including deep neural networks, random forests, bagging, and other stochastic optimization algorithms.
    In contrast, some learning procedures are deterministic once the data are fixed. Examples include ordinary least squares, logistic regression, kernel ridge regression, and other convex empirical-risk minimization methods solved by deterministic optimization routines. In these cases, the algorithmic randomness may be viewed as a degenerate random element, i.e., $R_q\equiv r_0 $ almost surely for some constant $r_0$. Consequently, the representation \eqref{eq:Tq-representation1} provides a unified framework that simultaneously covers both randomized and deterministic learners.
    
\end{remark}

\begin{assumpA}[Algorithmic Randomness]\label{assump:algorithmic-randomness}
The algorithmic random elements $R_1,\ldots,R_Q$ are i.i.d. and independent of all other sources of randomness used in the procedure.
\end{assumpA}

Next, we study the dependence among the statistics $\{T_1, T_2, \ldots, T_Q\}$ as well as the $p$-values $\{p_1, p_2, \ldots, p_Q\}$. Characterizing this dependence is essential for establishing the validity of IAD-C2ST and EIAD-C2ST. 
Note that the training of classifier $\widehat{f}_q$ relies on the same $\calD_X^{tr}$ but different $\calD_{Y,q}^{tr}$ for each $q \in \{ 1, \ldots, Q \}$, and the construction of $T_q$ relies on the same $\calD_X^{te}$ and $\calD_Y^{te}$ for all $q$. Consequently, the  statistics $\{T_1, T_2, \ldots, T_Q\}$ and hence the corresponding $p$-values $\{p_1, p_2, \ldots, p_Q\}$, are inherently dependent. Their dependence, therefore, requires careful investigation. 

\begin{theorem}[Exchangeability]
\label{thm:exchangeablity} 
With Assumption \ref{assump:uniform-assignment}--\ref{assump:algorithmic-randomness}, the statistics $\{T_1, \ldots, T_{Q_{n,m}}\}$ are exchangeable, i.e., for any permutation $\pi$ of $\{1, \ldots, Q_{n,m}\}$, we have $(T_1, \ldots, T_{Q_{n,m}}) \overset{d}{=} (T_{\pi(1)}, \ldots, T_{\pi(Q_{n,m})})$. As a result, the $p$-values $\{p_1, \ldots, p_{Q_{n,m}}\}$ are also exchangeable.
\end{theorem}

Theorem \ref{thm:exchangeablity} establishes the key justification for the (E)IAD-C2ST aggregation step: the blockwise $p$-values are exchangeable, and therefore the exchangeable harmonic-mean rule $g_{\rm exch}$ in \eqref{pval:HMP-exch-rule} yields a valid merged $p$-value under $H_0$, with the result of asymptotic size for IAD-C2ST 
formalized in Theorem \ref{thm:size-IAD-C2ST}.
Theorem \ref{thm:power-IAD-C2ST} further studies its asymptotic power. 

\begin{assumpA}[Null Prediction Nondegeneracy]
\label{assump:null-nondegeneracy}
Under $H_0:\mathcal P_X=\mathcal P_Y$, there exists a constant
$\underline\pi\in(0,1/2)$ such that, for all sufficiently large $(n,m)$,
\[
\mathbb P_{H_0}
\left(
\underline\pi
\le
\mathcal E_{1,\widehat f_q}
\le
1-\underline\pi
\quad
\text{for every }q\in[Q_{n,m}]
\right)
=1.
\]
\end{assumpA}
 
\begin{theorem}[Asymptotic Size of IAD-C2ST] \label{thm:size-IAD-C2ST}
With Assumptions \ref{assump:uniform-assignment}, \ref{assump:algorithmic-randomness}, \ref{assump:null-nondegeneracy} and Condition~(\ref{cond:sample-splitting-ratio}) and suppose $\log Q_{n,m}=o(n_{\min}^{1/3} )$, for any $\alpha \in (0,1)$, 
\begin{equation}\label{eq:thm-IAD-C2ST-size}
    \limsup_{\min\{n,m\} \rightarrow \infty} \bbP_{H_0} (p_{\rm IAD-C2ST} \leq \alpha) \leq \alpha,
\end{equation}
where $\bbP_{H_0}(\cdot ) = \bbP(\cdot \mid H_0)$ denotes the probability under the null hypothesis $H_0$.
\end{theorem}

Since the power of a classifier two-sample test depends on the ability of its learned classifier, we assume that a
blockwise learned classifier preserves the population classification
performance of a target classifier. The following condition formalizes
this requirement.

\begin{assumpA}[Population Error Approximation of the Learned Classifier]
\label{assump:learning-algorithm}
Let $f$ denote a population target classifier and let $\widehat f_q$
denote a classifier learned from a generic blockwise training sample.
There exists a deterministic sequence $r_{n,m}\to0$ such that
$
\max_{a\in\{-1,1\}}
\left|
\mathcal E_{a,\widehat f_q}
-
\mathcal E_{a,f}
\right|
=
O_{\mathbb P}(r_{n,m})
$
uniformly over the alternative parameter space.
For local alternatives satisfying
$
\Delta_f
:=
\frac12-\mathcal E_f
\ge\delta_{n,m},
$
we additionally assume
$
r_{n,m}=o(\delta_{n,m}).
$
\end{assumpA}

\begin{theorem}[Asymptotic Power of IAD-C2ST]
\label{thm:power-IAD-C2ST}
Let $\delta_{n,m}>0$ characterize the classification signal under the
alternative, and define
\[
\Theta_{\delta_{n,m}}
:=
\left\{
(\mathcal P_{\mathcal X},\mathcal P_{\mathcal Y}):
\mathcal P_{\mathcal X}\neq\mathcal P_{\mathcal Y},
\quad
\Delta_f
:=
\frac12-\mathcal E_f
\ge\delta_{n,m}
\right\}.
\]

Under Assumptions~\ref{assump:uniform-assignment},\ref{assump:algorithmic-randomness},\ref{assump:learning-algorithm} and Condition~(\ref{cond:sample-splitting-ratio}) and suppose $\log Q_{n,m}=o(n_{\min}^{1/3} )$,
the following statements hold: 
\begin{enumerate}[label=(\roman*)]

\item \textnormal{(Rate-aware inheritance of blockwise power consistency.)} 
Let
$
C_Q:=\log Q+\log\log Q+1
$
for $Q\geq 2$. 
As $\min\{n,m\}\to\infty$, 
if 
$
\min_{1\le q\le Q}
\left\{
(qC_Q+1)p_q
\right\}
\overset{\mathbb P}{\rightarrow}0,
$
then
$
p_{\mathrm{IAD-C2ST}}
\overset{\mathbb P}{\rightarrow}0.
$

\item \textnormal{(Uniform power consistency.)}  When $\delta_{n,m}$ satisfies the condition that 
\begin{equation}\label{cond:power-growing-Q}
    n_{\min}\delta_{n,m}^2- \log(C_Q+1)
\rightarrow\infty,
\end{equation} 
then, for every fixed
$\alpha\in(0,1)$,
\[
\inf_{\theta\in\Theta_{\delta_{n,m}}}
\mathbb P_\theta
\left(
p_{\mathrm{IAD-C2ST}}\le\alpha
\right)
\rightarrow 1
\qquad
\text{as }\min\{n,m\}\to\infty.
\]In particular, if $\delta_{n,m}\equiv\delta$ for some constant
$\delta>0$, then \eqref{cond:power-growing-Q} is automatically satisfied under
$\log Q_{n,m}=o(n_{\min}^{1/3})$. Hence, IAD-C2ST is uniformly consistent
over $\Theta_{\delta}$. 

\end{enumerate}
\end{theorem}
\begin{remark}[Fixed and diverging numbers of blocks]
\label{rem:power-fixed-growing-Q} 
Theorem~\ref{thm:power-IAD-C2ST} covers both fixed and diverging
sequences $Q=Q_{n,m}$.
(a) If $Q$ is fixed, then $C_Q$ is a fixed constant.
Consequently, for any fixed $q_0\in[Q]$, if 
$
p_{q_0}\overset{\mathbb P}{\rightarrow}0$, then
$ p_{\mathrm{IAD-C2ST}}
\overset{\mathbb P}{\rightarrow}0.$ Thus, in the fixed-$Q$ regime, IAD-C2ST inherits consistency from any split. Moreover, 
Equation~\eqref{cond:power-growing-Q} reduces to
$
\sqrt{n_{\min}}\delta_{n,m}\rightarrow\infty
$ for achieving consistent power. 
(b) If $Q$ diverges to $\infty$, ordinary convergence
$p_{q_{n,m}}\overset{\mathbb P}{\to}0$ is no longer sufficient because
the aggregation multiplier diverges. The blockwise $p$-value must instead
vanish fast enough to satisfy
$
(q_{n,m}C_Q+1)p_{q_{n,m}}
\overset{\mathbb P}{\rightarrow}0.
$
Since $C_Q\asymp\log Q$, a sufficient uniform signal condition is
\eqref{cond:power-growing-Q}, 
where
$
\log(C_Q+1)\asymp\log\log Q.
$
Thus, relative to the fixed-$Q$ regime, diverging $Q$ introduces only
a logarithmic-in-$\log Q$ aggregation penalty.
\end{remark}

The IAD-C2ST construction is asymmetric because it assigns one half of the data to training and the other half to testing. Under our stratified equal-split scheme, however, the two halves are generated symmetrically, and the labels ``training'' and ``testing'' are arbitrary before the algorithm is applied. Thus, for a realized split $\mathcal S$, both $(\calD^{tr}, \calD^{te})$ and $(\calD^{te}, \calD^{tr})$ are two equally legitimate splits. 
These two splits test the same null hypothesis but may yield different $p$-values in finite samples because classifier training and evaluation are not symmetric operations. Below, we study the dependence structure of $p$-values resulting from applying IAD-C2ST to the original and swapped train--test assignments. 

\begin{proposition}[Exchangeability of the train--test orientations]
\label{prop:orientation-exchangeability-EIAD-C2ST}
Suppose that the data are partitioned using a uniform stratified
equal-half split. Applying the same IAD-C2ST procedure to the
original and swapped train--test assignments, then,  with Assumption \ref{assump:algorithmic-randomness}, 
$p_{\rm IAD-C2ST}^{(1)}$ and $p_{\rm IAD-C2ST}^{(2)}$ are exchangeable, i.e., 
$(p_{\rm IAD-C2ST}^{(1)},
        p_{\rm IAD-C2ST}^{(2)})
    \overset{d}{=}
(p_{\rm IAD-C2ST}^{(2)},
        p_{\rm IAD-C2ST}^{(1)}).
$
This conclusion holds at every sample size for which the
specified split is defined, under both the null and alternative
hypotheses.
\end{proposition}

EIAD-C2ST is designed to remove the artificial train-test orientation dependence. Below, in Theorem \ref{thm:size-power-EIAD-C2ST}, we prove that EIAD-C2ST preserves asymptotic type-I error control under the null hypothesis, while it inherits power consistency under alternatives whenever either orientation-specific IAD-C2ST $p$-value converges to zero. 
It may improve finite-sample power when the initially chosen orientation is less favorable than the swapped orientation. Again, building on top of IAD-C2ST, EIAD-C2ST does not require the balanced-sample condition (\ref{cond:sample-size-ratio}), yielding strong performance in scenarios with highly imbalanced samples. 

\begin{theorem}[Asymptotic Size and Power of EIAD-C2ST] \label{thm:size-power-EIAD-C2ST}
With Assumptions \ref{assump:uniform-assignment}--\ref{assump:learning-algorithm} and Condition~(\ref{cond:sample-splitting-ratio}), and suppose $\log Q_{n,m}=o(n_{\min}^{1/3} )$, then, 
\begin{enumerate}
    \item[(i)] (Asymptotic size.) For any $\alpha \in (0,1)$, 
$$\limsup_{\min\{n,m\} \rightarrow \infty} \bbP_{H_0} (p_{\rm EIAD-C2ST} \leq \alpha) \leq \alpha.$$  
\item[(ii)] (Asymptotic power.) EIAD-C2ST inherits power consistency from IAD-C2ST:  
as $\min\{n,m\}\to\infty$, if $ p_{\rm IAD\text{-}C2ST} \overset{\mathbb P}{\to} 0$, then $ p_{\rm EIAD\text{-}C2ST} \overset{\mathbb P}{\to} 0$. 
Furthermore, when $\delta_{n,m}$ satisfies the condition \eqref{cond:power-growing-Q}, for any $\alpha\in(0,1)$,
    \[
    \inf_{\theta\in\Theta_{\delta_{n,m}}}
    \mathbb P_\theta
    \left(
    p_{\rm EIAD\text{-}C2ST}\leq \alpha
    \right)
    \to 1 \quad \text{ as } \min\{n,m\}\to\infty.
    \]
\end{enumerate}
\end{theorem}

Next, we study the theoretical properties of (E)IAD-C2ST under representation amortization. The representation amortization procedure changes how each fold classifier depends on the data. Previously, the training procedure $\widehat{f}_q=\texttt{Train}(\calD_X^{tr},\calD_{Y,q}^{tr})$ makes the dependence of $\widehat{f}_q$ on the majority training sample fold-local: besides the shared minority sample $\calD_X^{tr}$, the only majority input to $\widehat{f}_q$ is the $q$-th slice $\calD_{Y,q}^{tr}$. However, in the representation-amortized implementation, the frozen encoder $\widehat{\phi}$ is trained on the entire   $\calD_Y^{tr}$, so every $\widehat{f}_q=\widehat{h}_q\circ\widehat{\phi}$ now depends on all of $\calD_Y^{tr}$ through the common component $\widehat{\phi}$. Each fold classifier therefore uses additional information beyond $\calD_{Y,q}^{tr}$. 

To account for this extra layer of dependence, we use $R_0$ to denote the algorithmic random element of the training procedure  $\texttt{TrainEnc}(\cdot)$, and $R_1,\ldots,R_Q$ to denote the algorithmic
random elements used by the $Q$ probing procedures. 
Then, the fitting procedure of the RA classifier can be represented using 
\begin{equation*}
    \widehat{\phi}:= \texttt{TrainEnc}(\calD^{tr}) 
    = \Xi_{\rm enc} (\mathcal D_X^{\rm tr},\mathcal D_Y^{\rm tr},R_0)
\end{equation*}
and 
\begin{equation*}
    \widehat{h}_q := \texttt{Probe}(\{\widehat{\phi}(\calD_X^{tr}),\widehat{\phi}(\calD_{Y,q}^{tr})\})
    = \Xi_{\rm probe}(\widehat{\phi}, \calD^{tr}_X, \calD^{tr}_{Y,q}, R_q).
\end{equation*}
Thus, the resulting test statistic can be written as
    $$T^{\text{RA}}_{q} := \Psi(\widehat{h}_q\circ\widehat{\phi}, \calD_X^{te}, \calD_Y^{te}).$$
    Equivalently, 
    \begin{equation}\label{eq:Tq-representation}
       T^{\text{RA}}_{q} := \calT_{\rm RA}(\calD^{tr}_X, \calD^{tr}_{Y}, \calD^{tr}_{Y,q}, \calD_X^{te}, \calD_Y^{te}, R_0, R_q)
    \end{equation}
where $\Xi_{\rm enc}$, $\Xi_{\rm probe}$, $\Psi$, $\calT_{\rm RA}$ are measurable functions that do not depend on the fold label $q$.

\begin{assumpA}[Algorithmic Randomness for the Representation-Amortized Implementation]\label{assump:algorithmic-randomness-ra}
The algorithmic random
elements $R_1,\ldots,R_Q$ are i.i.d., $R_0$ is independent of
$(R_1,\ldots,R_Q)$, and
$(R_0,R_1,\ldots,R_Q)$
is independent of all other sources of randomness used in the procedure. 
\end{assumpA}

\begin{theorem}[Exchangeability under representation amortization]
\label{thm:exchangeablity-ra}
 Under Assumption
\ref{assump:uniform-assignment} and Assumption \ref{assump:algorithmic-randomness-ra},
the test statistics $\{T^{\text{RA}}_{ 1},\ldots,T^{\text{RA}}_{Q}\}$ are exchangeable. Consequently, the corresponding $p$-values
$\{p^{\text{RA}}_{1},\ldots,p^{\text{RA}}_{Q}\}$ are also exchangeable.
\end{theorem}

Theorem~\ref{thm:exchangeablity-ra} shows that the representation-amortized implementation preserves the exchangeability structure underlying the validity of (E)IAD-C2ST. Consequently, the asymptotic size and power guarantees established for (E)IAD-C2ST under full refitting implementation extend to their representation-amortized implementation. We formalize this result below.

\begin{theorem}[Asymptotic Size and Power of (E)IAD-C2ST with Representation Amortization]
\label{thm:size-power-RA-EIAD-C2ST}
Under Assumptions~\ref{assump:uniform-assignment}, \ref{assump:null-nondegeneracy}, \ref{assump:learning-algorithm}, \ref{assump:algorithmic-randomness-ra}, Condition~\eqref{cond:sample-splitting-ratio} as well as $\log Q_{n,m}=o(n_{\min}^{1/3})$, suppose that the tests are
constructed as described in Section~\ref{sec:scalable-EIAD-C2ST}.
Then the asymptotic size and power guarantees established for
(E)IAD-C2ST in Theorems~\ref{thm:size-IAD-C2ST},
\ref{thm:power-IAD-C2ST}, and
\ref{thm:size-power-EIAD-C2ST} also hold with representation amortization.
\end{theorem}

\section{Simulation Studies}\label{sec: sim}

In this section, we evaluate the finite-sample performance of (E)IAD-C2ST on two synthetic settings spanning both Euclidean and non-Euclidean data. 
Section~\ref{subsec:simu-Gaussian-data} considers a simple setting with Euclidean data.
Section \ref{subsec:simu-graph} studies a synthetic graph setting designed to assess the algorithm's behavior under structured non-Euclidean data with a topological signal.

To assess the empirical performance of the proposed (E)IAD-C2ST procedures, we compare them against the following baselines:
\begin{itemize}
\item[$-$] Vanilla C2ST (V-C2ST). The classifier is trained directly using the standard objective associated with the chosen backbone, without any explicit correction for class imbalance.

\item[$-$] Weighted C2ST (W-C2ST). The classifier is trained with a cost-sensitive objective function, in which class-balanced weights are used to account for unequal sample sizes.

\item[$-$] Oversampling C2ST (OS-C2ST). For Euclidean data, we use the Synthetic Minority Over-sampling Technique (SMOTE) \citep{chawla2002smote}. For non-Euclidean data, where SMOTE is not directly applicable, we use simple random minority oversampling by duplicating minority-class training samples until the two classes are balanced.

\item[$-$] Maximum Mean Discrepancy (MMD). For Euclidean data, we use the standard MMD test with a Gaussian radial basis function kernel \citep{gretton2012kernel}. For graph-valued data, we use graph-specific MMD baselines, i.e., Graph-MMD \citep{nikolentzos2021graph,gudmundarson2024gtst}.
The hyperparameter settings of MMD/Graph-MMD used in the simulation studies are reported in Appendix~\ref{append:mmd}. 
\end{itemize}

Across all C2ST-based methods, we use a stratified random train-test split, with 50\% of the observations assigned to training the classifier and the remaining 50\% assigned to constructing the test statistics.
We use (E)IAD-C2ST to refer to the proposed testing procedure irrespective of how the classifier representation is fitted. In settings where representation learning is inexpensive, EIAD-C2ST may be implemented by refitting the complete classifier across the inner splits. For representation-intensive classifiers, we use the representation-amortized implementation. The latter changes the computational realization of EIAD-C2ST but not the overarching testing construction. We therefore retain the common name (E)IAD-C2ST throughout, and explicitly identify the implementation only where relevant.

\subsection{Simulation Results for Euclidean Data}\label{subsec:simu-Gaussian-data}

We first consider a simple Euclidean setting in which the two samples are generated from a $p$-dimensional multivariate $t$ distribution, denoted by $t_{\nu,p} (\bmu, \bS)$, where $\nu$ is the degree of freedom, $\bmu$ is the location vector, and $\bS$ is the scale matrix. 
Specifically, the two populations are generated as
$\mathcal D_X=\{X_i\}_{i=1}^{n}\stackrel{\mathrm{i.i.d.}}{\sim} t_{3,p}(\bmu_1,\bS)$ and $\mathcal D_Y=\{Y_j\}_{j=1}^{m}\stackrel{\mathrm{i.i.d.}}{\sim} t_{3,p}(\bmu_2,\bS)$, where we keep the convention $n\leq m$ to make $\calD_X$ representing the minority sample and $\calD_Y$ for the majority sample. 
We set $\nu = 3$ and $p=200$ throughout, and let the two populations share the same dependence structure. We consider $\bSigma_\rho = \bigl(\rho^{|i-j|}\bigr)_{1\le i,j\le p}$, an autoregressive matrix with correlation parameter $\rho$. The scale matrix is adjusted via $\bS = (\nu-2)/\nu \cdot \bSigma_{\rho}$ so that the covariance of the multivariate Student-$t$ distribution is exactly $\bSigma_{\rho}$. 

For the backbone classifier, we use an $\ell_2$-penalized logistic classifier and assign class labels according to a probability threshold of $0.5$. 
For size evaluation, we set $\bmu_1=\bmu_2=\bzero$, corresponding to the null hypothesis $H_0$. 
For power evaluation, we center the majority population at zero by holding $\bmu_2=\bzero$, while introducing a mean shift in a subset of the coordinates, specifically, by setting the first $k$ coordinates of $ \bmu_1$ to have location $\delta$, while the remaining $p-k$ coordinates have location zero. Here, $\delta>0$ controls the signal strength, and $k$ is the number of non-zero coordinates that controls the sparsity of the alternative: smaller values of $k$ correspond to more localized and sparser differences between the two populations, whereas a larger $k$ represents a denser mean shift. In the power experiments, we set $\delta=0.1$ and vary $k$ in \{20,50,100\} for different signal sparsity patterns.
We conduct experiments to evaluate test performance under varying degrees of sample imbalance by setting $p = 200$, $n=100$ and varying $m\in\{400,900,1900,4900,9900,19900\}$. We also examine different correlation structures by taking $\rho\in\{0, 0.2,0.5,0.8\}$, representing uncorrelated, weakly correlated, moderately correlated, and highly correlated settings, respectively. %

\begin{table}[tb]
\centering
\resizebox{\textwidth}{!}{
\begin{tabular}{rrrr|cccccc}
\toprule
$n$ & $m$ & Ratio & $\rho$ & MMD
& V-C2ST & W-C2ST & OS-C2ST & IAD-C2ST & EIAD-C2ST\\
\midrule
100 & 19900 & 199 & 0.8 & 0.051 & 0.000 & 0.030 & 0.026 & 0.078 & 0.074 \\
100 &  9900 &  99 & 0.8 & 0.056 & 0.000 & 0.026 & 0.029 & 0.072 & 0.070 \\ 
100 &  4900 &  49 & 0.8 & 0.046 & 0.007 & 0.030 & 0.047 & 0.063 & 0.056 \\ 
100 &  1900 &  19 & 0.8 & 0.037 & 0.017 & 0.045 & 0.041 & 0.057 & 0.042 \\ 
100 &   900 &   9 & 0.8 & 0.043 & 0.028 & 0.040 & 0.048 & 0.040 & 0.034 \\ 
100 &   400 &   4 & 0.8 & 0.047 & 0.038 & 0.057 & 0.061 & 0.051 & 0.038 \\ 
\midrule
100 & 19900 & 199 & 0.5 & 0.046 & 0.000 & 0.032 & 0.029 & 0.071 & 0.072 \\ 
100 &  9900 &  99 & 0.5 & 0.050 & 0.001 & 0.022 & 0.027 & 0.067 & 0.059 \\ 
100 &  4900 &  49 & 0.5 & 0.040 & 0.009 & 0.030 & 0.039 & 0.067 & 0.059 \\ 
100 &  1900 &  19 & 0.5 & 0.036 & 0.028 & 0.032 & 0.036 & 0.049 & 0.035 \\ 
100 &   900 &  9  & 0.5 & 0.039 & 0.038 & 0.043 & 0.040 & 0.047 & 0.033 \\ 
100 &   400 &  4  & 0.5 & 0.051 & 0.056 & 0.053 & 0.044 & 0.042 & 0.018 \\ 
\midrule
100 & 19900 & 199 & 0.2 & 0.047 & 0.000 & 0.032 & 0.027 & 0.074 & 0.065 \\ 
100 &  9900 &  99 & 0.2 & 0.051 & 0.001 & 0.022 & 0.027 & 0.074 & 0.069 \\ 
100 &  4900 &  49 & 0.2 & 0.040 & 0.008 & 0.032 & 0.036 & 0.064 & 0.047 \\ 
100 &  1900 &  19 & 0.2 & 0.036 & 0.026 & 0.037 & 0.034 & 0.041 & 0.038 \\ 
100 &   900 &  9  & 0.2 & 0.042 & 0.043 & 0.045 & 0.044 & 0.044 & 0.033 \\ 
100 &   400 &  4  & 0.2 & 0.054 & 0.048 & 0.054 & 0.050 & 0.041 & 0.027 \\ 
\midrule
100 & 19900 & 199 & 0 & 0.047 & 0.000 & 0.031 & 0.025 & 0.076 & 0.068 \\ 
100 &  9900 &  99 & 0  & 0.050 & 0.001 & 0.022 & 0.029 & 0.069 & 0.065 \\ 
100 &  4900 &  49 & 0  & 0.039 & 0.007 & 0.033 & 0.039 & 0.056 & 0.050 \\ 
100 &  1900 &  19 & 0 & 0.039 & 0.022 & 0.034 & 0.034 & 0.049 & 0.047 \\ 
100 &   900 &  9  & 0 & 0.042 & 0.040 & 0.046 & 0.048 & 0.055 & 0.047 \\ 
100 &   400 &  4  & 0 & 0.054 & 0.056 & 0.047 & 0.046 & 0.034 & 0.028 \\ 
\bottomrule
\end{tabular}
}
\caption{Empirical size for the Euclidean simulation at the significance level $\alpha=0.05$. Results are reported based on $1,000$ Monte Carlo replications.} 
\label{tab:euclidean_mean_size_log_t3}
\end{table}

\begin{figure}[tb]
    \centering
    \includegraphics[width=0.8\linewidth]{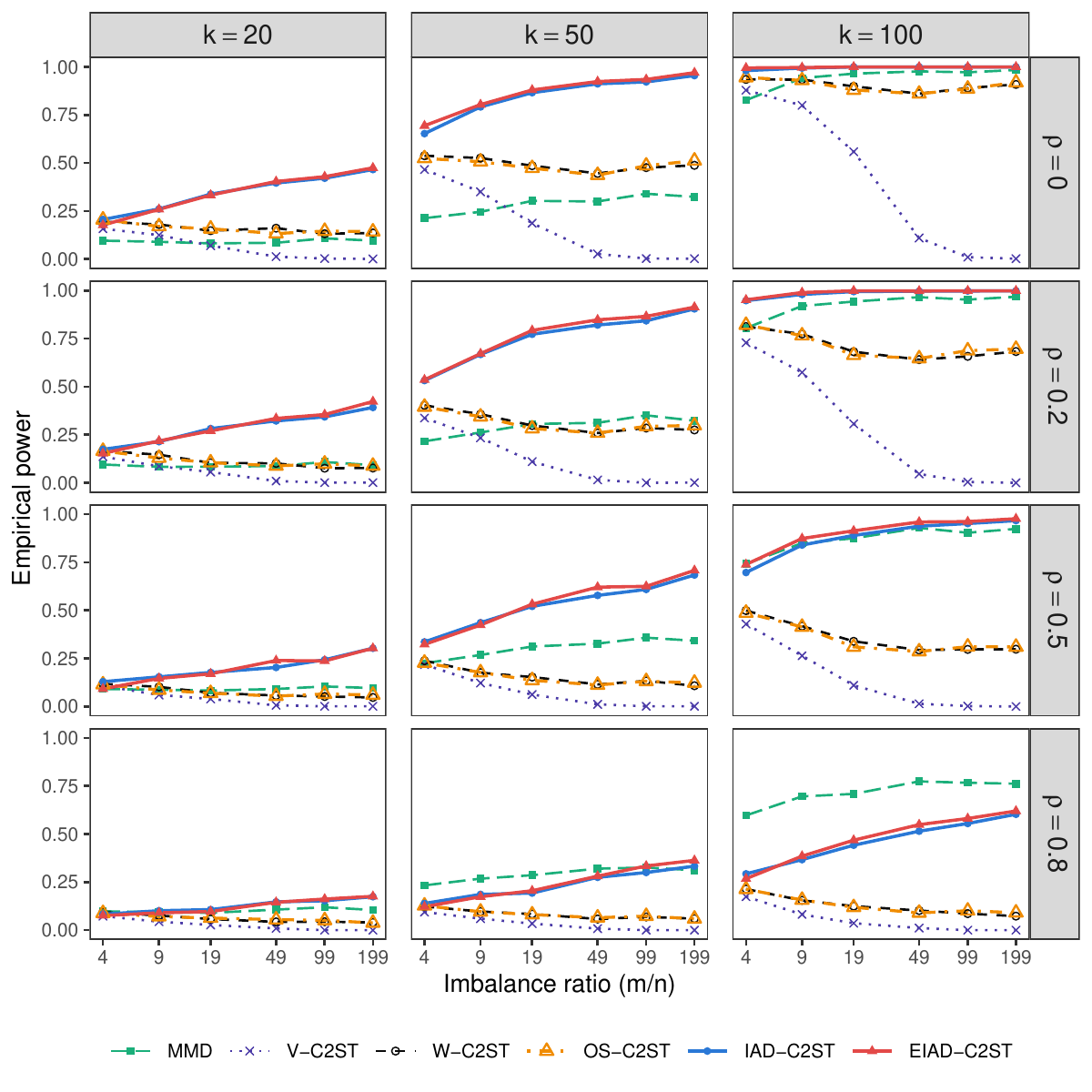}
    \vspace{-2ex}
    \caption{Empirical power for the Euclidean simulation at the significance level $\alpha=0.05$. Results are based on $1,000$ Monte Carlo replications.}
    \label{fig:euclidean_mean_power_log_t}
\end{figure}

Table~\ref{tab:euclidean_mean_size_log_t3} summarizes the empirical size results. Under the null, MMD remains well calibrated, as expected from its permutation-based implementation. Among the classifier-based procedures, 
V-C2ST becomes increasingly conservative as the sample imbalance grows, confirming the central motivation for our work: standard classifier two-sample testing can deteriorate substantially under severe class imbalance. 
W-C2ST and OS-C2ST partially alleviate this issue but remain somewhat conservative in many of the severely imbalanced settings. IAD-C2ST tends to be mildly liberal under the most extreme imbalance, whereas EIAD-C2ST provides noticeably better overall size calibration and stays closer to the nominal level across the simulation settings.

The power results in Figure~\ref{fig:euclidean_mean_power_log_t} demonstrate the main advantage of the proposed procedures. IAD-C2ST and EIAD-C2ST substantially outperform V-C2ST, W-C2ST, and OS-C2ST over a broad range of imbalance ratios and signal configurations, with the gains becoming particularly pronounced under severe sample imbalance. Notably, while the performance of V-C2ST can deteriorate as the majority sample grows, the proposed procedures are able to exploit the additional majority observations and often become more powerful as the imbalance increases. This supports the central motivation of (E)IAD-C2ST: partitioning the majority sample into balanced training subsets allows information in a large majority sample to be utilized without allowing class imbalance to dominate classifier training.

The advantage of the proposed procedures is most evident under weak-to-moderate feature dependence. As the within-feature correlation becomes strong, MMD becomes increasingly competitive and can outperform the classifier-based procedures for some relatively dense alternatives. Nevertheless, IAD-C2ST and EIAD-C2ST generally remain substantially more powerful than the other C2ST-based imbalance corrections. The two proposed procedures exhibit similar power overall, with EIAD-C2ST often providing modest gains while achieving better null calibration. Taken together, these results indicate that the proposed methods are particularly effective in the highly imbalanced regimes for which they were designed, while remaining competitive across a broad range of dependence and signal structures.

\subsection{Simulation Results for Graph Data}
\label{subsec:simu-graph}

We then consider a non-Euclidean setting in which the observations are graph-valued.
We generate graph two-sample datasets using a variant of the Stochastic Block Model (SBM) cluster regime \citep{dwivedi2023benchmarking}, designed to differentiate between positive (illicit) and negative (licit) graphs that differ only through a localized signal. 

Each graph is first generated from a base SBM with $K$ communities. The community sizes are sampled independently and uniformly on the integers $\{c_{\min}, \ldots, c_{\max}\}$. 
 The model is
parameterized by a within-community edge probability $p_{\mathrm{in}}$ and a baseline
between-community edge probability $q_{\mathrm{out}} < p_{\mathrm{in}}$, so that edges are
more likely within a community than across communities.
To vary the difficulty of the
base community structure, a mixing parameter $h \in [0,1]$ shrinks this gap via
$q_{\mathrm{out}}^{(h)}
\;:=\;
q_{\mathrm{out}} + 0.95\,h\,(p_{\mathrm{in}}-q_{\mathrm{out}})$, 
so that a larger $h$ moves the between-community probability toward $p_{\mathrm{in}}$
while preserving $q_{\mathrm{out}}^{(h)} < p_{\mathrm{in}}$, and $h = 0$ recovers
$q_{\mathrm{out}}$. 
Edges are then sampled independently, with within-community probability $p_{\mathrm{in}}$ and between-community probability $q_{\mathrm{out}}^{(h)}$. 

Licit graphs are left unchanged after this base generation step. For illicit graphs, we select one community uniformly at random and mark a fraction $\eta$ of its nodes as illicit, rounding down and always marking at least one node.
To make the class distinction learnable while preserving the “cluster” flavor, we apply two illicit-only perturbations, each controlled by a signal-strength parameter $s\geq 0$:  
(i) Intra-illicit densification: missing edges among illicit nodes are added with probability $s\lambda_{\mathrm{intra}}$, and 
(ii) Heterophily bridges: missing edges between illicit nodes and nodes outside the selected community are added with probability $s\lambda_{\mathrm{het}}$. 
Here, $\lambda_{\mathrm{intra}}$ and $\lambda_{\mathrm{het}}\in[0,1]$ are given constants that determine the base rates of the two perturbations. When $s=0$, licit and illicit graphs are generated from the same distribution, making the two classes structurally indistinguishable. This case corresponds to the null hypothesis and is used for size evaluation. As $s$ increases, the illicit graphs contain a stronger but still localized structural signal.
In our experiments, we use $K=6$, $(c_{\min},c_{\max})=(5,35)$, $(p_{\mathrm{in}},q_{\mathrm{out}})=(0.55,0.25)$, $\eta=0.40$, and $(\lambda_{\mathrm{intra}},\lambda_{\mathrm{het}})=(0.3,\,0.1)$.  

For the synthetic experiments on graph-valued data, we use a GNN with a GIN backbone and
attention graph pooling, with the graph topology as input. Here, GIN refers to the Graph Isomorphism Network, a message-passing
GNN architecture introduced by \citet{xu2018how}. 
For (E)IAD-C2ST with GNN classifiers, we use our representation-amortized implementation to reduce the computational cost of repeatedly training the GNNs.
Details of the GNN architecture, hyperparameter settings, and classifier-training procedures
are provided in Appendix~\ref{append:graph sim2}.

For each Monte Carlo replication, we generate $m$ licit graphs and $n$ illicit graphs using the data-generating process described above. 
The signal strength $ s$ governs the departure from the null hypothesis: $ s = 0$ corresponds
to the null, under which we assess size, and $ s > 0$ to alternatives, under which we assess power.
We consider different values of the mixing parameter $h$, which controls community separability.

\begin{table}
\centering
\resizebox{\textwidth}{!}{
\begin{tabular}{crrrrr|cccccc}
\toprule
 Type& $n$ & $m$   &Ratio& \multicolumn{1}{c}{Signal($s$)} & \multicolumn{1}{c|}{Mixing($h$)} & Graph-MMD& V-C2ST & W-C2ST & OS-C2ST & IAD-C2ST &EIAD-C2ST\\
\midrule
size  & 100 & 9900  & 99 & 0 & 0 & 0.053 & 0.000 & 0.081 & 0.020 & 0.067 & 0.066 \\ 
      & 100 & 4900  & 49 & 0 & 0 & 0.047 & 0.000 & 0.053 & 0.025 & 0.041 & 0.042 \\ 
      & 100 & 1900  & 19 & 0 & 0 & 0.056 & 0.000 & 0.092 & 0.028 & 0.040 & 0.042 \\ 
      & 100 &  900  &  9 & 0 & 0 & 0.043 & 0.000 & 0.064 & 0.016 & 0.041 & 0.038 \\ 
      & 100 &  400  &  4 & 0 & 0 & 0.056 & 0.000 & 0.056 & 0.022 & 0.042 & 0.038 \\ 
      & 100 &  200  &  2 & 0 & 0 & 0.053 & 0.006 & 0.048 & 0.033 & 0.039 & 0.033 \\ 
      \cmidrule{2-12}
      & 100 & 9900  & 99 & 0 & 0.3 & 0.052 & 0.000 & 0.063 & 0.037 & 0.067 & 0.067 \\  
      & 100 & 4900  & 49 & 0 & 0.3 & 0.052 & 0.000 & 0.054 & 0.026 & 0.054 & 0.055 \\ 
      & 100 & 1900  & 19 & 0 & 0.3 & 0.054 & 0.000 & 0.080 & 0.029 & 0.054 & 0.038 \\ 
      & 100 &  900  &  9 & 0 & 0.3 & 0.044 & 0.000 & 0.063 & 0.021 & 0.038 & 0.033 \\ 
      & 100 &  400  &  4 & 0 & 0.3 & 0.061 & 0.000 & 0.043 & 0.030 & 0.039 & 0.027 \\  
      & 100 &  200  &  2 & 0 & 0.3 & 0.062 & 0.008 & 0.033 & 0.034 & 0.042 & 0.033 \\ 
      \cmidrule{2-12}
      & 100 & 9900  & 99 & 0 & 0.4 & 0.050 & 0.000 & 0.067 & 0.037 & 0.065 & 0.064 \\ 
      & 100 & 4900  & 49 & 0 & 0.4 & 0.050 & 0.000 & 0.061 & 0.042 & 0.043 & 0.041 \\ 
      & 100 & 1900  & 19 & 0 & 0.4 & 0.053 & 0.000 & 0.086 & 0.025 & 0.046 & 0.050 \\ 
      & 100 &  900  &  9 & 0 & 0.4 & 0.047 & 0.000 & 0.070 & 0.018 & 0.044 & 0.040 \\ 
      & 100 &  400  &  4 & 0 & 0.4 & 0.063 & 0.000 & 0.050 & 0.028 & 0.031 & 0.034 \\ 
      & 100 &  200  &  2 & 0 & 0.4 & 0.055 & 0.006 & 0.047 & 0.028 & 0.047 & 0.037 \\ 
      \midrule      
power & 100 & 9900 & 99 & 1.5 & 0.3 & 0.214 & 0.001 & 0.620 & 0.637 & 0.974 & \textbf{0.992} \\ 
      & 100 & 4900 & 49 & 1.5 & 0.3 & 0.205 & 0.000 & 0.626 & 0.545 & 0.984 & \textbf{0.992} \\ 
      & 100 & 1900 & 19 & 1.5 & 0.3 & 0.185 & 0.000 & 0.341 & 0.649 & 0.808 & \textbf{0.916} \\ 
      & 100 &  900 &  9 & 1.5 & 0.3 & 0.189 & 0.003 & 0.410 & 0.275 & 0.849 & \textbf{0.954} \\ 
      & 100 &  400 &  4 & 1.5 & 0.3 & 0.165 & 0.005 & 0.279 & 0.209 & 0.790 & \textbf{0.903} \\ 
      & 100 &  200 &  2 & 1.5 & 0.3 & 0.136 & 0.112 & 0.222 & 0.207 & 0.607 & \textbf{0.715} \\ 
       \cmidrule{2-12}
      & 100 & 9900 & 99 & 1.5 & 0.4 & 0.177 & 0.000 & 0.470 & 0.586 & 0.933 & \textbf{0.972} \\ 
      & 100 & 4900 & 49 & 1.5 & 0.4 & 0.162 & 0.000 & 0.497 & 0.456 & 0.940 & \textbf{0.974} \\ 
      & 100 & 1900 & 19 & 1.5 & 0.4 & 0.164 & 0.000 & 0.289 & 0.563 & 0.754 & \textbf{0.894} \\ 
      & 100 & 900  &  9 & 1.5 & 0.4 & 0.128 & 0.001 & 0.345 & 0.220 & 0.796 & \textbf{0.915} \\ 
      & 100 & 400  &  4 & 1.5 & 0.4 & 0.153 & 0.005 & 0.209 & 0.166 & 0.722 & \textbf{0.819} \\ 
      & 100 & 200  &  2 & 1.5 & 0.4 & 0.131 & 0.063 & 0.193 & 0.171 & 0.515 & \textbf{0.625} \\ 
      \cmidrule{2-12}
      & 100 & 9900 & 99 & 2.0 & 0.3 & 0.315 & 0.052 & 0.950 & 0.909 & \textbf{1.000} & \textbf{1.000} \\ 
      & 100 & 4900 & 49 & 2.0 & 0.3 & 0.308 & 0.031 & 0.940 & 0.789 & \textbf{1.000} & \textbf{1.000} \\ 
      & 100 & 1900 & 19 & 2.0 & 0.3 & 0.262 & 0.106 & 0.655 & 0.900 & 0.960 & \textbf{0.996} \\ 
      & 100 &  900 &  9 & 2.0 & 0.3 & 0.283 & 0.108 & 0.740 & 0.659 & 0.961 & \textbf{0.999} \\ 
      & 100 &  400 &  4 & 2.0 & 0.3 & 0.224 & 0.058 & 0.558 & 0.496 & 0.942 & \textbf{0.995} \\ 
      & 100 &  200 &  2 & 2.0 & 0.3 & 0.197 & 0.246 & 0.404 & 0.408 & 0.814 & \textbf{0.938} \\ 
\bottomrule
\end{tabular}
}
\caption{Empirical size and power for the Graph simulation under varying imbalance and signal settings. Tests are conducted at $\alpha=0.05$, and results are based on $1,000$ Monte Carlo replications. The highest power in each row is shown in \textbf{bold}.
}
\label{tab:cluster_graph_power}
\end{table}

Table~\ref{tab:cluster_graph_power} reports the results for graph-valued data. The size results are broadly consistent with those observed in the Euclidean setting. Graph-MMD remains well calibrated under the null, as expected from its permutation-based implementation, while V-C2ST becomes highly conservative under severe imbalance. Weighting and oversampling alleviate this issue to some extent, although their size can deviate from the nominal level in several settings. IAD-C2ST and EIAD-C2ST maintain reasonably stable size across the reported graph configurations.

The power results show a clear advantage of the proposed procedures.
Both IAD-C2ST and EIAD-C2ST substantially outperform V-C2ST, W-C2ST, OS-C2ST, and Graph-MMD across the considered imbalance, signal-strength, and mixing settings. The gains are particularly substantial under severe sample imbalance, where V-C2ST has very little power and W-C2ST and OS-C2ST partially alleviate this issue. In this experiment, the class difference is characterized by a localized topological perturbation embedded within an SBM-generated graph rather than a Euclidean shift.  Graph-MMD provides a natural nonparametric benchmark, but it exhibits relatively low power for the topological structural alternatives considered here. IAD-C2ST and EIAD-C2ST detect these differences substantially more effectively than the alternative imbalance corrections and Graph-MMD across the considered configurations. The gains remain pronounced even under severe imbalance, indicating that the proposed strategy can effectively combine a flexible GNN representation with balanced classifier training to extract information from a large majority sample. 

The graph design further illustrates the roles of signal strength and structural complexity. Increasing $s$ strengthens the illicit topological signal and consequently increases power, whereas increasing the mixing parameter $h$ weakens community separation and makes the localized perturbation more difficult to detect. The proposed procedures remain effective as this structural separation deteriorates. Moreover, EIAD-C2ST consistently matches or improves upon IAD-C2ST across the reported alternatives, with the benefit of symmetrization being more apparent here than in the Euclidean experiments.

The graph setting is particularly relevant to the motivation of our work. For Euclidean data, widely used tools such as SMOTE provide a direct mechanism for addressing class imbalance, while kernel-based procedures such as MMD offer an established alternative to classifier-based testing. Such options are considerably more limited for complex non-Euclidean observations. In particular, SMOTE cannot be directly applied to graph-valued data without defining a meaningful interpolation operation on the graph space. Although graph-specific MMD procedures provide a natural nonparametric benchmark, they exhibit relatively low power for the localized topological alternatives considered here. In contrast, the proposed framework addresses imbalance at the level of classifier training and therefore does not require synthetic interpolation of the original observations.

Overall, the results highlight a practical advantage of the proposed (E)IAD-C2ST for complex non-Euclidean data: it accommodates flexible deep classifiers while directly addressing severe sample imbalance, particularly in settings where standard Euclidean remedies are either not readily applicable or may offer limited power.

\section{Application to Blockchain Forensics on Bitcoin Transaction Graphs}\label{sec: Real Dataset}

We apply the proposed methods to a real-world blockchain forensics problem using the Elliptic2 dataset \citep{bellei2024shape}. Elliptic2 is a large-scale, graph-structured dataset of Bitcoin transactions, in which each node represents a cluster of Bitcoin addresses believed to be controlled by the same entity, and each directed edge represents a transaction between two address clusters. The dataset is publicly available at \url{https://github.com/MITIBMxGraph/Elliptic2}.

\begin{figure}[!tb]
    \centering
    \includegraphics[width=0.8\linewidth]{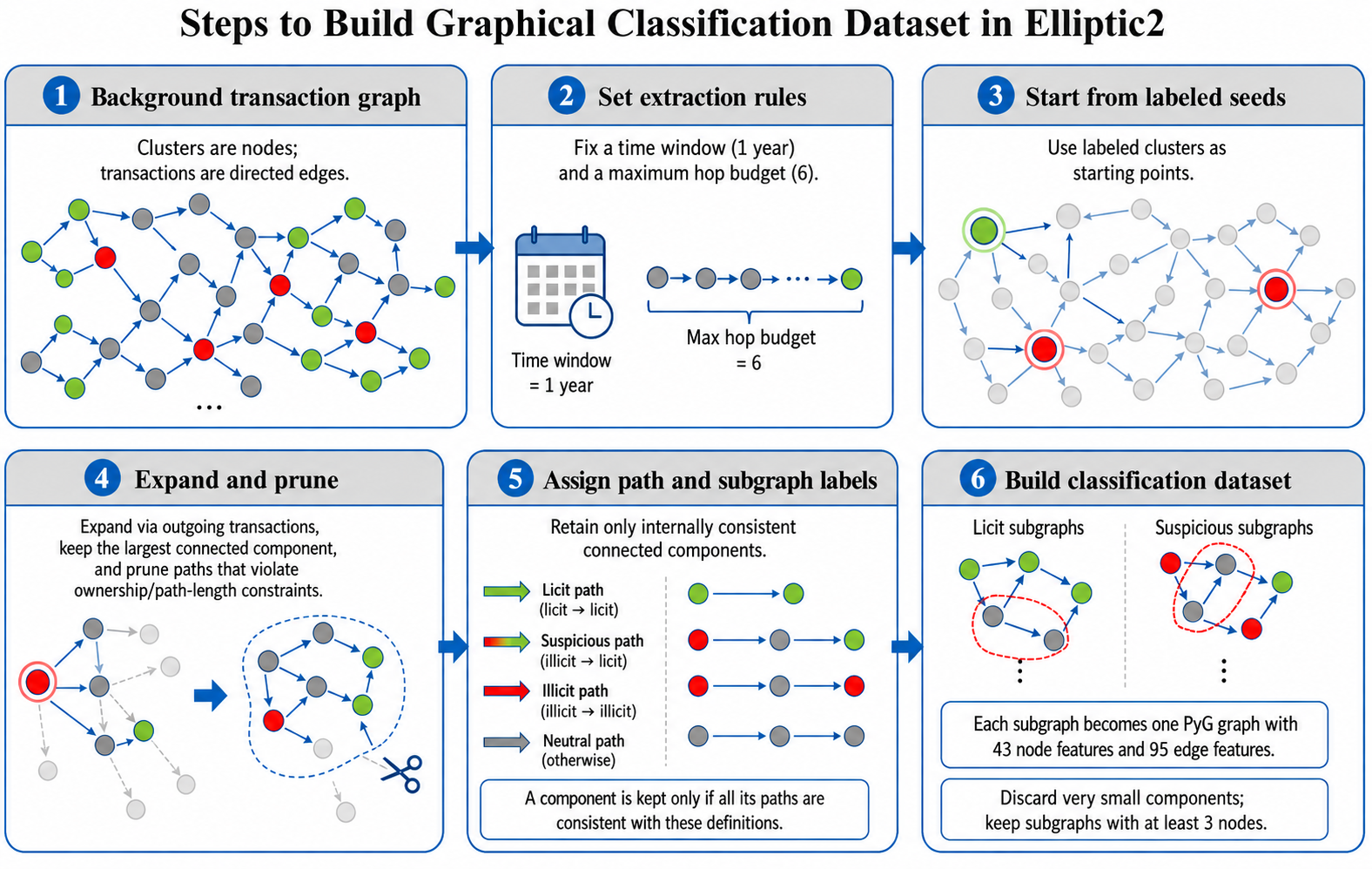}
    \caption{Schematic illustration of the construction of labeled subgraphs
    from the Elliptic2 Bitcoin transaction graph.}
    \label{fig:Graph_Construction}
\end{figure}

The full background graph contains 49{,}299{,}864 nodes and 196{,}215{,}606 directed edges. We follow the guidelines developed by \citet{bellei2024shape}, detailed below, to construct and annotate the subgraphs. 
Each subgraph represents a local flow of Bitcoin transactions among address clusters and is labeled based on the identities of the clusters at the endpoints of its constituent transaction paths. 
A schematic illustration of the construction procedure is presented in  Figure~\ref{fig:Graph_Construction}.

The construction begins by specifying a time window, a maximum traversal
depth, and an early-stopping criterion indicating a likely change in
ownership of the funds. The largest connected component within the
selected time window is used as the background graph. Starting from
labeled clusters, the graph is traversed along outgoing transactions until
another labeled cluster is reached, the hop limit is exceeded, or the
early-stopping criterion is satisfied. The number of outgoing transactions
used to initiate traversal is capped for each starting cluster to limit
the influence of differences in transaction activity. For the released
Elliptic2 dataset, the time window covers one year, and the maximum
traversal depth is set to six hops.
The extracted paths are classified according to their endpoint labels.
These path labels are then used to identify components that admit a
consistent subgraph annotation. 

To describe the subgraph annotations more precisely, we first define the cluster and path types used in Elliptic2. 

\begin{definition} 
\label{def: cluster}
    A \textit{licit cluster} is owned by a “licit” entity (exchange, wallet provider, miner, licit service, etc.). An \textit{illicit cluster} is owned by an “illicit” entity (dark market, scam, hack, etc.). Clusters that are neither licit nor illicit are deemed \textit{unknown} (e.g., unlabeled clusters).
\end{definition}

\begin{definition}
\label{def: path}
    Each path found in the extraction step is labeled by its two endpoint clusters. A \textit{licit path} connects a licit cluster to a licit cluster; a \textit{suspicious path} connects an illicit cluster to a licit cluster; an \textit{illicit path} connects an illicit cluster to an illicit cluster. Any other path, for example one ending at an unknown cluster, is considered \textit{neutral}.
\end{definition}

\begin{figure}[!tb]
    \centering
    \includegraphics[width=0.7\linewidth]{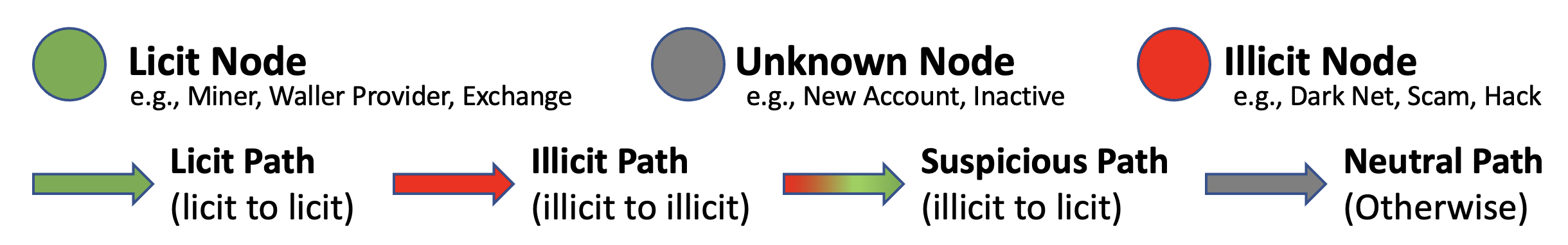}
    \vspace{-2ex}
    \caption{\small Illustration of different cluster and path types used in the subgraph construction.}
    \label{fig:nodes and paths}
\end{figure}

\begin{figure}[!tb]
    \centering
    \includegraphics[width=0.7\linewidth]{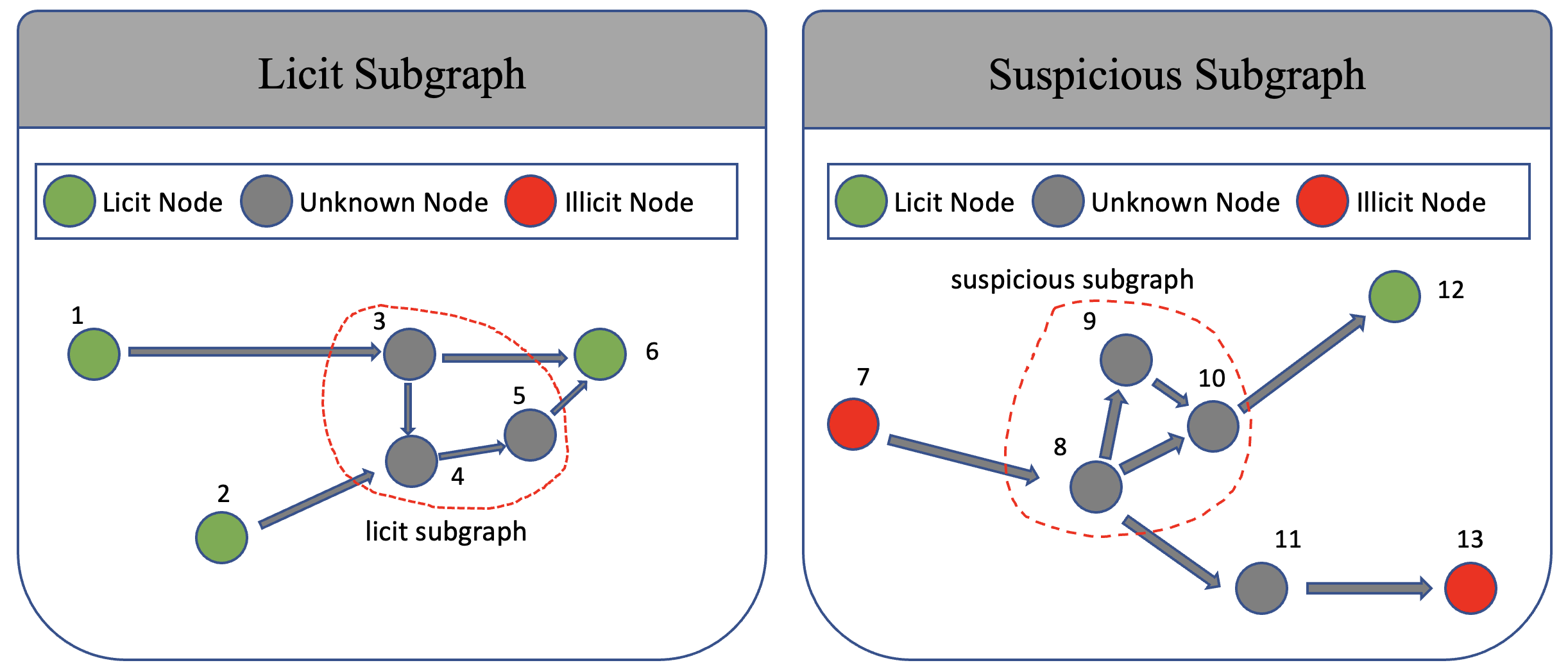}
    \vspace{-2ex}
    \caption{\small Illustration of licit and suspicious subgraph construction.
The left example contains three licit paths:
$1\to3\to6$,
$1\to3\to4\to5\to6$, and
$2\to4\to5\to6$.
The right example contains two suspicious paths:
$7\to8\to10\to12$ and
$7\to8\to9\to10\to12$,
together with one illicit path:
$7\to8\to11\to13$.
The red dashed boundaries enclose the unknown nodes retained in the
final annotations: $\{3,4,5\}$ for the licit subgraph and
$\{8,9,10\}$ for the suspicious subgraph.
Node $8$ is retained because it lies on suspicious paths, whereas
node $11$ is excluded because it lies only on the illicit path.
Labeled endpoint nodes are excluded from both annotated subgraphs.}
    \label{fig:Subgraph_Example}
\end{figure}

Candidate components obtained from the extracted paths are checked for label consistency. A component can yield a licit annotation when it contains a licit path and its paths are restricted to licit and neutral types. Similarly, a component can yield a suspicious annotation when it contains a suspicious path and its paths are restricted to suspicious, illicit, and neutral types. Components containing incompatible path types, such as both licit and suspicious paths, are left unlabeled and excluded from the binary classification task.
For each retained component, the final subgraph annotation includes only unknown clusters that lie on paths associated with the assigned class. Thus, unknown clusters on licit paths form a \textit{licit} subgraph, whereas unknown clusters on suspicious paths form a \textit{suspicious} subgraph. A node remains eligible for inclusion when it lies on a class-defining path even if it also belongs to an illicit or neutral path. By contrast, nodes lying exclusively on illicit or neutral paths are excluded from the final annotation.
Consequently, \textit{neutral} and \textit{illicit} are path labels used during construction, not additional classes in the final subgraph classification task. The released task distinguishes only licit and suspicious subgraphs. The known endpoint clusters support the annotation procedure but are excluded from the final annotated node sets; the released dataset provides subgraph labels rather than the cluster-level labels used during construction. 
An illustrative example of the path labeling and final subgraph annotation is provided in Figure~\ref{fig:Subgraph_Example}.

A defining feature of this application is the substantial class imbalance. From the background graph, 121{,}810 labeled transaction subgraphs are constructed for binary classification into \textit{licit} and \textit{suspicious} classes.  Among them, 119{,}047 are \textit{licit}, whereas only 2{,}763 are \textit{suspicious}. The licit subgraphs contain an average of 3.65 nodes, with sizes ranging from 2 to 296 nodes, while the suspicious subgraphs contain an average of 3.79 nodes, with sizes ranging from 2 to 30 nodes. Most features in the released dataset have been discretized into ordinal bins, and we represent each labeled subgraph as a separate PyTorch Geometric (PyG) graph. To focus on subgraphs with nontrivial graph structure, we retain only those containing at least three nodes. After filtering, the dataset contains 62{,}071 labeled subgraphs, consisting of 60{,}546 licit and 1{,}525 suspicious subgraphs, or approximately $39.7$ licit subgraphs for every suspicious subgraph. This combination of complex graph-valued observations and severe class imbalance provides a natural real-world setting for evaluating the proposed framework.

For the classifier backbone, we use a GNN based on the Graph Isomorphism Network with Edge features (\textsc{GINE}) \citep{Hu2020Strategies}, together with attention-based graph pooling. Because repeatedly fitting a full GNN becomes computationally demanding under severe imbalance, we employ the representation-amortized implementation of (E)IAD-C2ST introduced in Section~\ref{sec:scalable-EIAD-C2ST}. Details of the network architecture, optimization settings, and classifier training procedure are provided in Appendix~\ref{append:realdata}.

As in the graph simulation study, we compare IAD-C2ST and EIAD-C2ST with V-C2ST, W-C2ST, OS-C2ST, and Graph-MMD. To evaluate empirical size, we construct a null setting in which both samples are drawn from the licit class, so that there is no systematic distributional difference between the two populations. To evaluate power, we use the licit and suspicious labels provided by the dataset, drawing $m$ subgraphs from the licit class and $n$ subgraphs from the suspicious class. This construction preserves the practically relevant setting in which suspicious transaction patterns constitute the minority population. %

\begin{table}[tb]
\centering
\resizebox{0.9\textwidth}{!}{
\begin{tabular}{crrr|cccccc}
\toprule
  Type& \multicolumn{1}{c}{\textbf{$n$}} & \multicolumn{1}{c}{\textbf{$m$}} & Ratio & Graph-MMD & V-C2ST & W-C2ST & OS-C2ST & 
IAD-C2ST &EIAD-C2ST \\
\midrule
size  & 100 &  9900  &  99 & 0.052 & 0.000 & 0.001 & 0.005 & 0.050 & 0.045 \\ 
      & 100 &  4900  &  49 & 0.066 & 0.000 & 0.008 & 0.010 & 0.038 & 0.035 \\ 
      & 100 &  1900  &  19 & 0.061 & 0.000 & 0.023 & 0.019 & 0.044 & 0.032 \\
      & 100 &   900  &   9 & 0.044 & 0.000 & 0.039 & 0.028 & 0.035 & 0.032 \\ 
      & 100 &   400  &   4 & 0.051 & 0.002 & 0.043 & 0.031 & 0.037 & 0.024 \\ 
      \cmidrule{2-10}
      & 200 &  9900 & 49.5 & 0.044 & 0.000 & 0.008 & 0.015 & 0.028 & 0.028 \\ 
      & 200 &  4900 & 24.5 & 0.046 & 0.000 & 0.026 & 0.028 & 0.028 & 0.024 \\
      & 200 &  1900 &  9.5 & 0.046 & 0.000 & 0.027 & 0.027 & 0.027 & 0.022 \\ 
      & 200 &   900 &  4.5 & 0.054 & 0.001 & 0.049 & 0.026 & 0.029 & 0.019 \\ 
      & 200 &   400 &    2 & 0.050 & 0.008 & 0.056 & 0.033 & 0.027 & 0.020 \\ 
      \midrule
power & 100 &  9900  &  99 & 0.388 & 0.016 & 0.392 & 0.525 & 0.791 & \textbf{0.912} \\
      & 100 &  4900  &  49 & 0.376 & 0.030 & 0.475 & 0.586 & 0.742 & \textbf{0.894} \\ 
      & 100 &  1900  &  19 & 0.369 & 0.035 & 0.427 & 0.615 & 0.634 & \textbf{0.767} \\ 
      & 100 &   900  &   9 & 0.343 & 0.014 & 0.390 & 0.494 & 0.484 & \textbf{0.556} \\ 
      & 100 &   400  &   4 & 0.297 & 0.160 & 0.382 & 0.304 & 0.408 & \textbf{0.444} \\ 
      \midrule 
      & 200 &  9900 & 49.5 & 0.811 & 0.225 & 0.819 & 0.949 & 0.988 & \textbf{1.000} \\ 
      & 200 &  4900 & 24.5 & 0.797 & 0.317 & 0.904 & 0.950 & 0.986 & \textbf{0.999} \\ 
      & 200 &  1900 &  9.5 & 0.753 & 0.475 & 0.793 & 0.933 & 0.930 & \textbf{0.985} \\ 
      & 200 &   900 &  4.5 & 0.687 & 0.440 & 0.792 & 0.844 & 0.868 & \textbf{0.935} \\ 
      & 200 &   400 &    2 & 0.556 & 0.355 & 0.623 & 0.521 & 0.702 & \textbf{0.813} \\ 
      \bottomrule
\end{tabular}
}
\caption{Empirical size and power of tests on subgraphs drawn from the Elliptic2 dataset. Tests are conducted at the significance level $\alpha = 0.05$, and results are based on $1,000$ Monte Carlo replications. The highest power in each row is shown in \textbf{bold}.}
\label{tab:elliptic_results}
\end{table}

Table~\ref{tab:elliptic_results} summarizes the results for the Elliptic2 application. The null experiment serves as a calibration check for the real-data setting. IAD-C2ST and EIAD-C2ST maintain rejection rates at or below the nominal level across the reported configurations, while Graph-MMD is also well calibrated through its permutation-based implementation. By contrast, V-C2ST is nearly degenerate under substantial imbalance, providing further evidence that direct classifier training is poorly suited to this setting without explicit imbalance correction.

The power results provide a clearer distinction among the methods. EIAD-C2ST achieves the highest power in every reported configuration, with IAD-C2ST also performing strongly relative to the competing procedures. The advantage is particularly pronounced when the suspicious sample is small and the imbalance is severe. As the size of the licit majority sample increases, the power of the proposed methods generally improves, indicating that they are able to exploit the additional majority observations rather than being adversely affected by the resulting class imbalance. Increasing the suspicious-sample size from \(n=100\) to \(n=200\) further improves detection power, with EIAD-C2ST approaching or reaching unit power in the more highly imbalanced settings.

These findings are especially relevant in the present application because the observations are complex transaction subgraphs rather than Euclidean feature vectors. Graph-MMD exhibits substantially lower power than the proposed procedures across the reported alternatives. The Elliptic2 results complement the synthetic graph experiments by showing that the advantages of (E)IAD-C2ST persist in a large-scale real-world network setting with genuine class imbalance and heterogeneous graph structure. In particular, the consistently strong performance of EIAD-C2ST provides empirical support for combining imbalance-aware training with the symmetrized testing construction in graph-based forensic applications.

\section{Conclusion and Discussion}\label{sec:discussion} 

This paper develops a new perspective for C2ST under severe sample imbalance.
We propose (E)IAD-C2ST, an imbalance-aware procedure that turns sample imbalance from a source of statistical difficulty into a usable source of information.
IAD-C2ST uses balanced blockwise fitting to generate multiple classifier-based tests, and combines their evidence with calibration that accounts for their dependence.
EIAD-C2ST removes the orientation choice and reduces the split-induced instability. The representation-amortized implementation makes the repeated-testing construction computationally practical while retaining its inferential structure under the corresponding assumptions.

A key feature of the proposed approach is that imbalance is handled through the construction of the test statistic rather than through conventional learning heuristics. EIAD-C2ST exploits the surplus majority sample by repeatedly forming approximately balanced classification subproblems and aggregating the resulting evidence. This construction departs fundamentally from conventional imbalance remedies such as loss reweighting or resampling \citep{chawla2002smote,fernandez2018smote}. This design separates the statistical role of imbalance from the optimization problem faced by the classifier and leads to a test that is both practically effective and broadly adaptable.

A central technical challenge is dependence: the subproblems share the same minority training sample and the same testing sample, so the resulting subproblem-level $p$-values are inherently dependent, placing the problem outside the reach of most of the $p$-value combination literature which operates on the (asymptotic) independence \citep{yu2023power,yu2024fisher,yu2024power}.
We resolve this challenge by theoretically proving that 
the blockwise $p$-values produced by the imbalance-aware construction are exchangeable, a property that may be of independent interest for other multi-split inference procedures.
Such exchangeability licenses valid aggregation through exchangeable merging rules \citep{vovk2020combining,gasparin2025combining}. We further show that (E)IAD-C2ST inherits consistent power from the subproblems under mild classification-power conditions. 
Empirically, across simulations and the real-world application involving various data types and backbone classifiers,  the proposed procedures consistently improve power in the severely imbalanced regimes that motivated this study, while maintaining proper Type-I error control. 

More broadly, this work suggests that severe imbalance should not be viewed merely as a nuisance for classifier training. When incorporated into the testing design, it can be exploited to construct more informative testing procedures. The imbalance-aware perspective developed here may contribute to a broader class of modern nonparametric tests involving complex, heterogeneous, and unevenly sampled data.
Our framework also suggests a path toward hypothesis testing in the ``long tail'' of machine learning \citep{liu2019large,zhang2023deep}, where rare classes, rare subpopulations, rare shifts, and rare failures often carry the greatest scientific and societal importance. 
Promising directions include rare failure mode detection in 
diverse safety-critical domains, such as jailbreak susceptibility in LLM systems \citep{wei2023jailbroken}, rare adverse events in clinical trials \citep{hong2021meta},  cybersecurity monitoring \citep{khraisat2019survey}, and the evaluation of algorithmic performance on clinical subgroups in medical studies \citep{buolamwini2018gender}.
In these applications, existing machine-learning approaches often focus primarily on point prediction, whereas the proposed imbalance-aware learning framework provides a pathway toward statistically principled inference for rare but consequential phenomena.
Beyond new hypothesis testing problems, it would also be of interest to investigate more efficient multi-split or cross-fitting variants that further reduce the variability induced by a single train-test split.
More generally, imbalance-aware testing with increasingly expressive deep classifiers also offers a promising route toward efficient and powerful hypothesis tests that can leverage the predictive capabilities of modern black-box machine learning algorithms for inference with rigorous statistical guarantees.  

\bibliographystyle{plainnat}
\bibliography{ref-DeepC2ST}

\end{document}